\documentclass[10pt,journal,compsoc]{IEEEtran}
\usepackage{amsmath,amsfonts}
\usepackage{algorithmic}
\usepackage{algorithm}
\usepackage{array}
\usepackage[caption=false,font=normalsize,labelfont=sf,textfont=sf]{subfig}
\usepackage{textcomp}
\usepackage{stfloats}
\usepackage{url}
\usepackage{verbatim}
\usepackage{graphicx}
\usepackage{cite}
\usepackage{subfiles}
\usepackage{booktabs}
\usepackage{xcolor}
\usepackage{multirow}
\usepackage{wrapfig}
\usepackage{colortbl}
\usepackage{makecell}
\usepackage{pifont}

\begin{document}



\title{EvolveNav:  Empowering LLM-Based Vision-Language Navigation via Self-Improving Embodied Reasoning}

\author{Bingqian~Lin\IEEEauthorrefmark{1}, Yunshuang Nie\IEEEauthorrefmark{1}, Khun Loun Zai, Ziming Wei, Mingfei Han, Rongtao Xu, Minzhe Niu, Jianhua Han, Hanwang Zhang, Liang Lin, Bokui Chen\IEEEauthorrefmark{2}, Cewu Lu\IEEEauthorrefmark{2}, Xiaodan Liang\IEEEauthorrefmark{2}
	\IEEEcompsocitemizethanks{
	\IEEEcompsocthanksitem 
	\IEEEauthorrefmark{1}These two authors contribute equally to this work.\protect\\
	\IEEEcompsocthanksitem 
	\IEEEauthorrefmark{2}Bokui Chen, Cewu Lu, and Xiaodan Liang are the corresponding authors.\protect\\
    \IEEEcompsocthanksitem Bingqian Lin and Cewu Lu are with Shanghai Jiao Tong University, Shanghai, China.
    \protect\\
    E-mail: \{linbq666, lucewu\}@sjtu.edu.cn
    \IEEEcompsocthanksitem 
    Yunshuang Nie, Khun Loun Zai, and Ziming Wei are with Shenzhen Campus of Sun Yat-sen University, Shenzhen, China. \protect\\
			E-mail:\{nieysh@mail2.sysu.edu.cn, khunlz@mail2.sysu.edu.cn, weizm3@mail2.sysu.edu.cn\}
            \IEEEcompsocthanksitem Xiaodan Liang is with Shenzhen Campus of Sun Yat-sen University, Shenzhen, China, Peng Cheng Laboratory, Guangdong Key Laboratory of Big Data Analysis and Processing, Guangzhou, 510006, China.
  \protect\\
  E-mail: liangxd9@mail.sysu.edu.cn

 \IEEEcompsocthanksitem Bokui Chen is with Tsinghua Shenzhen International Graduate School, Tsinghua University, China. \protect\\
 E-mail: chenbk@tinghua.edu.cn. 
  
		\IEEEcompsocthanksitem Mingfei Han is with the Department of Computer Vision, Mohamed Bin Zayed University of Artificial Intelligence, Abu Dhabi, UAE. \protect\\
 E-mail: mingfei.han@mbzuai.ac.ae.

 \IEEEcompsocthanksitem Rongtao Xu is with the Department of Computer Vision, Mohamed Bin Zayed University of Artificial Intelligence, Abu Dhabi, UAE, and also with Spatialtemporal AI. \protect\\
 E-mail: xurongtao2022@gmail.com.
 \IEEEcompsocthanksitem Minzhe Niu and Jianhua Han are with Yinwang Intelligent Technology Co., Ltd. \protect\\
 E-mail: hanjianhua4@huawei.com, niuminzhe1@huawei.com.

\IEEEcompsocthanksitem Hanwang Zhang is with the School of Computer Science and Engineering, Nanyang Technological University, Singapore. \protect\\
 E-mail:  hanwangzhang@gmail.com.

 \IEEEcompsocthanksitem Liang Lin is with Sun Yat-sen University, Guangzhou, China. \protect\\
 E-mail: linlng@mail.sysu.edu.cn.

	}
		}

\markboth{Journal of \LaTeX\ Class Files,~Vol.~14, No.~8, August~2021}%
{Shell \MakeLowercase{\textit{et al.}}: A Sample Article Using IEEEtran.cls for IEEE Journals}

\IEEEpubid{0000--0000/00\$00.00~\copyright~2021 IEEE}

\maketitle

\begin{abstract}
 Recent studies have revealed the potential of training open-source Large Language Models (LLMs) to unleash LLMs' reasoning ability for enhancing  vision-language navigation (VLN) performance, and simultaneously mitigate the domain gap between LLMs' training corpus and the VLN task. However, these approaches predominantly adopt straightforward input-output mapping paradigms, causing the mapping learning difficult and the navigational decisions unexplainable. Chain-of-Thought (CoT) training is a promising way to improve both navigational decision accuracy and interpretability, while the complexity of the navigation task makes the perfect CoT labels unavailable and may lead to overfitting through pure CoT supervised fine-tuning. 
 To address these issues, we propose \textbf{EvolveNav}, a novel  s\textbf{E}lf-impro\textbf{v}ing emb\textbf{o}died reasoning paradigm
that realizes adaptable and generalizable navigational reasoning for boosting \textbf{L}LM-based \textbf{v}ision-languag\textbf{e} \textbf{Nav}igation. Specifically,  EvolveNav involves a two-stage training process: (1) Formalized CoT Supervised Fine-Tuning, where we train the model with curated  formalized CoT labels to first activate the model's navigational reasoning capabilities, and simultaneously increase the reasoning speed;  
 (2) Self-Reflective Post-Training, where the model is iteratively trained with its own reasoning outputs as self-enriched CoT labels to enhance the supervision diversity. 
 A self-reflective auxiliary task is also designed to encourage the model to learn correct reasoning patterns by contrasting with wrong ones.
 Experimental results under both task-specific and cross-task training paradigms demonstrate the consistent superiority of EvolveNav over previous LLM-based VLN approaches on various popular benchmarks, including R2R, REVERIE, CVDN, and SOON. EvolveNav open avenues for exploring effective self-improving reasoning paradigms, enabling building agents capable of self-evolving for promoting LLM-based embodied AI research.
\end{abstract}

\section{Introduction} 
Vision-Language Navigation (VLN) has received significant research interest within the Embodied AI community, due to its practicality and flexibility in enabling human-robot interaction in real-world robotic applications. In VLN tasks, an embodied agent needs to follow natural language instructions to navigate through complex visual environments to reach the target position. 
Early works improve VLN performance by designing dedicated model architectures~\cite{wang2019reinforced,ma2019self, deng2020evolving,qi2020Object}, introducing powerful learning paradigms~\cite{zhu2020vision,li2019robust,huang2019transferable}, and developing useful data augmentation techniques~\cite{tan2019learning, fried2018speaker,liu2021vision,Fu2019CounterfactualVN}. Subsequently, pretraining-based VLN approaches have been widely proposed to improve the cross-modal alignment ability and decision accuracy of navigation agents~\cite{hong2021vln,Chen2021HistoryAM,Qi2021TheRT,Chen2022ThinkGA,Qiao2022HOPHA}. Nevertheless, constrained by the limited scale of pretraining and VLN in-domain data, these approaches cannot learn navigational reasoning knowledge 
sufficiently, and therefore still struggle to 
handle various unseen navigation scenarios.  

\begin{figure*}
\begin{centering}
\includegraphics[width=0.98\linewidth]{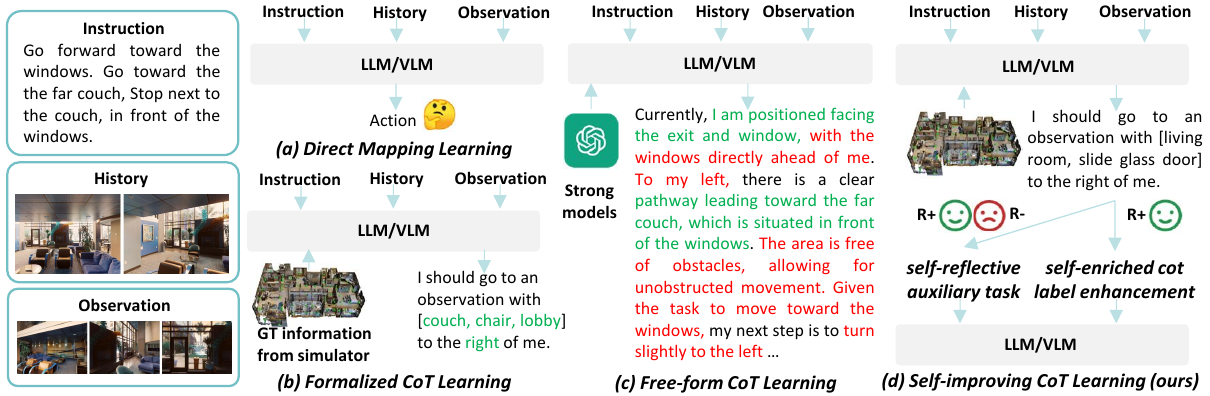}
\par\end{centering}
\caption{Comparison of different chain-of-thought (CoT) training paradigms. (a) Direct Mapping Learning maps the navigation inputs to actions straightforwardly. (b) Formalized CoT Learning and (c) Free-form CoT Learning generate formalized and free-form reasoning, respectively, under the training with fixed CoT labels. (d) Different from the above paradigms, our Self-Improving CoT Learning framework utilizes the model's own reasoning outputs as self-enriched CoT labels and learn the reasoning in a self-reflective way during CoT training to fulfill generalizable and adaptable reasoning. \textcolor{red}{Red} and \textcolor{green}{green} fonts represent wrong and correct reasoning outputs, respectively. \textit{R+} and \textit{R-} represent positive and negative reasoning samples, respectively.
}
\label{fig:motivation}
\end{figure*}

With the rapid progress of large language models (LLMs)~\cite{brown2020language,touvron2023llama,touvron2023llama2}, emerging works have introduced LLMs to address embodied tasks by resorting to LLMs' rich real-world common sense and powerful reasoning ability~\cite{Ahn2022DoAI,Huang2022InnerME,driess2023palme}. 
Some recent works have attempted to build LLM-based VLN models in a zero-shot or trainable manner
~\cite{chen2024mapgpt,zhou2023navgpt,long2023discuss,hao2020towards,zhang2024navid}. The zero-shot approaches, such as NavGPT~\cite{zhou2023navgpt} and MapGPT~\cite{chen2024mapgpt}, resort to closed-source LLMs~\cite{OpenAI_2023} 
to generate navigational reasoning and decision for different navigation timesteps.
To alleviate the high cost of frequently querying closed-source LLMs for sequential decision making,
the trainable methods 
collect in-domain data to train open-source LLMs~\cite{vicuna} 
to build the navigation agent. However, they typically map directly from navigational inputs to decisions without explicit intermediate reasoning steps, leading to decision uninterpretability and may also limit performance (see Figure~\ref{fig:motivation}(a)). 

Recent studies have revealed the effectiveness of  chain-of-thoughts (CoT) training in enhancing both the decision accuracy and interpretability for embodied tasks~\cite{zawalski2024robotic,zhao2025cot,lin2024navcot}.
Most of these approaches employ the supervised fine-tuning (SFT) paradigm for conducting embodied CoT training. 
However, 
introducing CoT supervised fine-tuning for training VLN models is highly challenging due to the following two reasons.
Firstly, due to the complexity and uncertainty of the navigation task, there can be multiple cues for deciding the correct navigation action, i.e., 
there may be no single ``correct'' CoT label to guide navigation for a specific timestep.
This leads to a hard collection process of perfect navigation CoT supervision. 
Secondly, pure CoT supervised fine-tuning using fixed CoT labels may cause overfitting to certain reasoning patterns and thus harm the generalization to diverse unseen scenarios.

In this paper, we propose a novel s\textbf{E}lf-impro\textbf{v}ing emb\textbf{o}died reasoning paradigm for enhancing \textbf{L}LM-based \textbf{v}ision-languag\textbf{e} \textbf{Nav}igation, called \textbf{EvolveNav}, to fulfill generalizable and adaptable navigational reasoning under 
various tasks and scenarios.
EvolveNav comprises two training phases: 1) {\it Formalized CoT Supervised Fine-Tuning} and 2) {\it Self-Reflective Post-Training}. 
The Stage 1 training of {\it Formalized CoT Supervised Fine-Tuning} aims to first activate the model's potential reasoning capabilities, where we ask the model to produce explicit chain-of-thought navigational reasoning dynamically by predicting the landmarks needed to locate with the corresponding direction for deciding the navigation actions. To alleviate generating redundant reasoning and increase the inference speed, we conduct the CoT supervised fine-tuning using curated formalized CoT labels, which are collected by filling the landmark and direction information into concise label templates. 
Then, we conduct {\it Self-Reflective Post-Training} for Stage 2 training, aiming to  mitigate the overfitting to pre-constructed CoT labels and enable self-improving reasoning for enhancing generalization.
Specifically, we design a self-enriched CoT label enhancement scheme, where we train the model with its iteratively produced correct reasoning outputs to diversify the CoT supervision. 
We also construct a self-reflective auxiliary task, where the model needs to discriminate between positive and negative navigational reasoning to learn correct reasoning patterns. 
As shown in Figure~\ref{fig:motivation}, in contrast to CoT supervised fine-tuning using fixed CoT labels (i.e., Figure~\ref{fig:motivation}(b) Formalized CoT Learning and (c) Free-form CoT Learning), our EvolveNav (Figure~\ref{fig:motivation}(d)) can generate embodied CoT in a self-refining manner during training to mitigate the overfitting. Additionally, through training with formalized CoT labels, our EvolveNav can significantly reduce uninformative navigational reasoning to promote the reasoning speed compared with using free-form CoT labels for training (as in Figure~\ref{fig:motivation} (c)). 



We conduct substantial experiments under both task-specific and cross-task training paradigms on multiple public VLN benchmarks, including R2R~\cite{anderson2018vision}, CVDN~\cite{thomason2019vision}, REVERIE~\cite{qi2020reverie}, and SOON~\cite{zhu2021soon}.
Experimental results show that EvolveNav 
significantly
outperforms previous LLM-based VLN approaches on various benchmarks, demonstrating the effectiveness of our self-improving embodied reasoning paradigm in promoting navigation decision accuracy and generalization.
We carefully conduct ablation experiments to explore how to design streamlined CoTs that can provide interpretability while boosting navigation performance. Visualization also insightfully reveals the reasonability of our design for CoT labels in enhancing decision interpretability and improving navigational reasoning.

To summarize, the main contributions of this paper are: 
\begin{itemize}

\item {We propose \textbf{EvolveNav}, a novel self-improving embodied reasoning paradigm  for enhancing LLM-based vision-and-language navigation, which fulfills generalizable and adaptable navigational reasoning under  various tasks and scenarios.  }



\item {We construct formalized CoT labels for conducting supervised fine-tuning, which effectively activates the agent's navigational reasoning ability and promotes the reasoning speed. We introduce a self-enriched CoT label enhancement strategy and a self-reflective auxiliary task to enable learning correct reasoning patterns in a self-refining manner to mitigate overfitting.}

\item{
Experimental results demonstrate the superiority of EvolveNav over previous LLM-based approaches on various VLN benchmarks. Our EvolveNav can improve the generalization of both navigational reasoning and decision-making, providing meaningful insights for designing advanced embodied reasoning paradigms. }

\end{itemize}

\section{Related Work}

\subsection{Vision-Language Navigation (VLN)}
Vision-Language Navigation (VLN) has attracted intensive research interest in recent years. 
Various VLN benchmarks have been proposed to evaluate agents' ability for navigational reasoning and instruction following~\cite{anderson2018vision,jain2019stay,ku2020room,thomason2019vision,qi2020reverie,zhu2021soon}. 
Previous approaches employ non-pretraining-based~\cite{wang2019reinforced,ma2019self, deng2020evolving,zhu2020vision,tan2019learning, fried2018speaker} or pretraining-based paradigms~\cite{hong2021vln,Chen2021HistoryAM,Chen2022ThinkGA,Qiao2022HOPHA,Guhur2021AirbertIP,an2022bevbert,wang2023scaling} for tackling the above VLN tasks. 
However, these approaches cannot generalize well to diverse unseen scenarios that require rich real-world commonsense, and the navigation decisions also lack explainability.
Some recent works have introduced LLMs to assist the VLN task, by either eliciting the useful navigation knowledge stored in LLMs~\cite{lin2024correctable,zhou2024navgpt,qiao2023march} or employing the LLM as the navigation backbone for action decision~\cite{zhou2023navgpt,zheng2024towards,zhang2024navid,lin2024navcot,chen2024mapgpt}. Our work lies in the latter. 

Different from previous approaches, we propose a new VLN framework in this work, where the LLM-based navigation backbone iteratively generates intermediate reasoning steps in a {\it self-improving} manner during training to guide navigational decisions. As a result, both the reasoning ability and decision interpretability of the navigation model can be significantly enhanced.


\subsection{LLMs as Embodied Agents}
Recent research have revealed  the giant potential of utilizing Large Language Models (LLMs) as embodied agents to complete the robotic navigation and manipulation tasks, benefiting from the outstanding ability of planning, reasoning, and reflection of LLMs~\cite{song2023llmplanner,Ahn2022DoAI,Huang2022InnerME,Yao2022ReActSR,shahlm,schumann-2023-velma,wang2023voyager,huang2023voxposer,rajvanshi2024saynav}. For example, LM-Nav~\cite{shahlm} introduces the LLM to parse the long navigation instruction into sequential landmarks for facilitating the navigational planning. Voxposer~\cite{huang2023voxposer} introduces LLMs for code writing and combines them with the Vision-Language models (VLMs) to compose 3D value maps for robotic manipulation. 

There are typically two branches of works where the LLMs act as embodied agents for tackling the VLN task. In the first branch,  closed-source LLMs like GPT-4~\cite{openai} are queried in a zero-shot manner to decide the action sequentially~\cite{zhou2023navgpt,long2023discuss,chen2024mapgpt,chen2024affordances}. For example, NavGPT~\cite{zhou2023navgpt} transforms visual observations into textual formats and feed them to LLMs for generating action predictions. The second branch finetunes open-source  LLMs with in-domain VLN datasets, which alleviates the LLM's query cost as well as mitigates the gap between LLM's training corpus and VLN tasks~\cite{zheng2024towards,lin2024navcot,zhang2024navid,han2024roomtour3d}. For example, Navid~\cite{zhang2024navid} constructs a video-based navigational vision-language model and train it using navigation samples collected from continuous R2R datasets. However, most of them map navigation inputs to action decisions directly without the
reasoning output. In contrast, we train the open-source LLM to generate self-improving embodied reasoning explicitly to improve action decision accuracy, which enhances the decision interpretability as well as mitigates overfitting to training reasoning labels.

\subsection{Embodied Chain-of-thoughts Training}
Chain-of-thoughts (CoT) reasoning~\cite{wei2022chain} has been a widely utilized technique in Large Language Models (LLMs) and Vision-Language Models (VLMs). By generating the intermediate reasoning steps rather than directly predicting the answer, CoT can promote the answer accuracy for various tasks such as mathematical reasoning, commonsense reasoning, code generation, {\it etc}~\cite{openai,OpenAI_2023,liu2024deepseek}. Inspired by this, some recent works have trained LLMs/VLMs to generate reasonable CoT for improving the action decision accuracy in embodied tasks~\cite{zawalski2024robotic,lin2024navcot,mu2023embodiedgpt,zhao2025cot,liu2025spatialcot,zhou2024navgpt}. ECoT~\cite{zawalski2024robotic} trains vision-language-action models to generate embodied reasoning including object bounding boxes and end effector positions to encourage better adaptation for robotic manipulation tasks. NavGPT-2~\cite{zhou2024navgpt} resorts to the GPT-4V model to collect free-form step-wise CoT reasoning data to improve the navigational reasoning ability of the VLN-specialized models~\cite{Chen2022ThinkGA}. CoT-VLA~\cite{zhao2025cot} incorporates explicit visual chain-of-thought reasoning into vision-language-action models by predicting future image frames before generating the action sequence. 

In this work, we propose a novel self-improving embodied CoT training paradigm for boosting  VLN performance, which effectively improves the reasoning ability and decision accuracy of the navigation model in various unseen scenarios. Moreover, we construct CoT supervision in a formalized manner, which significantly reduces redundant reasoning information and simultaneously promotes the model's inference speed.  


\begin{figure*}
\begin{centering}
\includegraphics[width=0.98\linewidth]{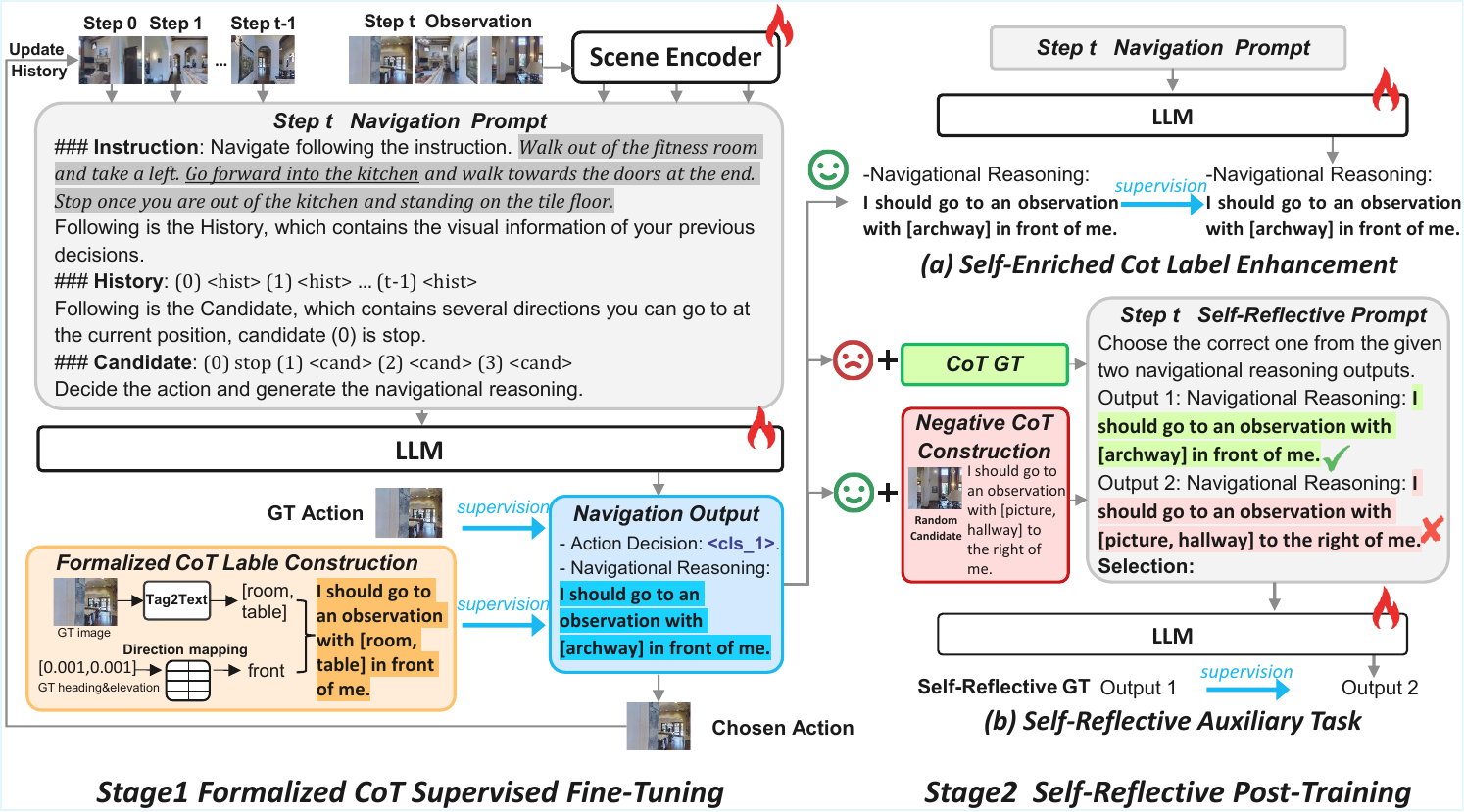}
\par\end{centering}
\caption{\textbf{Overview of EvolveNav}. EvolveNav involves a two-phase training framework for fulfilling self-improving embodied reasoning. In {\it Stage 1 Formalized CoT Supervised Fine-Tuning}, the navigation agent is trained using pre-constructed formalized CoT labels to generate navigational reasoning by predicting the landmark needed to locate with the corresponding direction. In {\it Stage 2 Self-Reflective Post-Training}, the agent's own reasoning outputs are introduced as the self-enriched CoT labels to enhance the supervision diversity. A self-reflective auxiliary task is also designed to guide the navigation agent to discriminate between correct and wrong reasoning outputs.
}
\label{fig:overview}
\end{figure*}

\section{Method}
\label{Method}
In this section, we first introduce the problem definition of the VLN task (Sec.~\ref{Problem Setup}). Then, we present the model architecture of the LLM-based navigation agent in our EvolveNav (Sec.~\ref{Model Architecture}). 
Finally, we delve into the details of the proposed self-improving embodied reasoning framework (Sec.~\ref{Self-Improving Embodied Reasoning}).

The overview of our EvolveNav is presented in Figure~\ref{fig:overview}. 
Specifically, EvolveNav consists of two training stages: 1) Formalized CoT Supervised Fine-Tuning (Sec.~\ref{Formalized CoT Supervised Fine-Tuning}), 
where we curate formalized CoT labels to initially train the LLM-based VLN model with supervised fine-tuning to activate the model's navigational reasoning ability and simultaneously promote the reasoning speed;
2) Self-Reflective Post-Training (Sec.~\ref{Self-Reflective Post Training}), 
where the model is further trained with its own reasoning outputs as self-enriched CoT labels to increase supervision diversity, accompanied by a self-reflective auxiliary task to encourage better learning of accurate navigational reasoning patterns by discriminating from incorrect ones.

\subsection{Problem Setup}
\label{Problem Setup}
In the VLN task, an agent is given a navigation instruction $I$ in the form of a declarative sentence or a dialogue and is required to navigate from a start position to the target position. 
At timestep $t$, the agent receives a panoramic observation $O_{t}$ containing $K$ single-view observations $O_{t,k}$, i.e., $O_{t}=\{O_{t,k}\}_{k=1}^{K}$.
There are $N$ navigable views 
among $K$ views. 
The 
navigable
views 
and the stop action form the action space, 
from which the agent chooses one as the action prediction $a_{t}$.
Actions before step $t$ are treated as the navigation history.

\subsection{Model Architecture}
\label{Model Architecture}
We build the LLM-based navigation agent that can simultaneously produce the navigational chain-of-thought reasoning and action prediction, modifying from a recent LLM-based VLN work, NaviLLM~\cite{zheng2024towards}. 
The navigation agent consists of a scene encoder $F_v$, an LLM backbone $F_{\mathrm{LLM}}$, and an action prediction head $F_{\mathrm{action}}$.
At each timestep $t$, the agent receives the navigation instruction $I$, panoramic observation $O_{t}$, and navigation history features $H_{t}=\{h_{0},...,h_{t-1}\}$. The scene encoder transforms $N$ navigable panoramic views $\{O_{t,n}\}_{n=1}^{N}$ into visual representations $\{V_{t,n}\}_{n=1}^{N}$:
\begin{align}
\{V_{t,n}\}_{n=1}^{N} &= F_{v}(\{O_{t,n}\}_{n=1}^{N}). 
\end{align}
The navigation prompt $P$ is then constructed by integrating the tokenized instruction, the visual representations $\{V_{t,n}\}_{n=1}^{N}$, and navigation history features $H_{t}$. As shown in Figure~\ref{fig:overview}, special tokens $<$hist$>$ and $<$cand$>$ are introduced as placeholder tokens, where we insert the features $H_{t}$ and $\{V_{t,n}\}_{n=1}^{N}$, respectively. 

In contrast to
NaviLLM~\cite{zheng2024towards} that directly maps the navigational inputs to action decision, in EvolveNav, we construct the following output hint in the prompt $P$ to guide the navigation agent to generate both the action decision and explicit navigational reasoning: ``-Action Decision: $<$cls$>$. -Navigational Reasoning: ''. The $<$cls$>$ token is also a special token for facilitating subsequent action predictions. The navigation prompt $P$ is fed into the LLM backbone $F_{\mathrm{LLM}}$ to obtain the feature ${f}^{cls}_{t}$ of the $<$cls$>$ token and the chain-of-thought (CoT) reasoning $\mathrm{CoT}$:
\begin{align}
{f}^{cls}_{t}, \mathrm{CoT} &= F_{\mathrm{LLM}}(P).
\end{align}
Under the guidance of CoT reasoning, the ${f}^{cls}_{t}$ is sent to the action prediction head $F_{\mathrm{action}}$ for generating the action prediction $a_{t}$:
\begin{align}
{a}_{t} &= F_{\mathrm{action}}({f}^{cls}_{t}).
\end{align}

\subsection{Self-Improving Embodied Reasoning Framework}
\label{Self-Improving Embodied Reasoning}

\subsubsection{Stage 1: Formalized CoT Supervised Fine-Tuning}
\label{Formalized CoT Supervised Fine-Tuning}
\noindent\textbf{Formalized CoT Labels Collection.}
When facing a given human instruction, 
the navigation agent usually needs to sequentially reason about the direction or the landmark it should move to in its current visual observation to reach the target position.
Therefore, in EvolveNav, we train the LLM-based VLN model to generate the chain-of-thought (CoT) reasoning about the landmark with the corresponding direction at different navigation timesteps, like the following format: ``{\it I should go to an observation with [landmark] to the [direction] of me}''.

To encourage the navigation agent to choose the ground-truth action $a^{*}_{t}$ (paired observation is denoted as $O_{t}^{*}$) at different timesteps $t$ through the guidance of CoT reasoning during training, we obtain the corresponding landmarks $L$ and direction $D$ of $O_{t}^{*}$ to construct the formalized CoT labels, which is described as follows. 
Denote the observation $O_{t}^{*}$ as $O_{t}^{*}=\{B_{t},A_{t}=\{\psi_{t},\theta_{t}\}\}$, where $B_{t}$ is the RGB image of $O_{t}^{*}$, $A_{t}$ represent the direction information containing heading $\psi_{t}$ and elevation $\theta_{t}$.
For the image $B_{t}$, we first employ a powerful image captioning model~\cite{huang2023tag2text} $F_{\mathrm{cap}}$ to obtain object and scene context $C_{t}$:
\begin{equation}
C_{t} = F_{\mathrm{cap}}(B_{t}).
\end{equation}
Then, we leverage the NLP tool Spacy~\cite{spacy} to extract the landmarks list $L$ from $C_{t}$. In contrast to directly using object recognition models which may detect multiple redundant objects, extracting landmarks from the image captions can better retain salient landmarks. As a result, the generated CoT reasoning of the navigation agent can effectively help it locate important landmarks mentioned in the human instruction, since humans also tend to focus on salient landmarks when giving navigation instructions. We follow~\cite{lin2024navcot} to map the direction information $A_{t}$ of the observation $O_{t}^{*}$ to textual represented direction $D$. With the landmarks list $L$ and direction  $D$, we construct CoT labels $\mathrm{\mathrm{CoT}^{*}}$ by filling the following label template: {\it I should go to an observation with [$L$] to the [$D$] of me}.

Through extracting the landmark and direction information of the ground-truth observation (action) straightforwardly to construct the CoT labels, we do not explicitly correlate the CoT labels and the navigation instructions, leading to excellent generalization to navigational instructions of various types. Such CoT label construction strategy can effectively alleviate the problem that some action decisions are not explicitly corresponding to the navigation instruction, while instead enabling latent alignment learning of action decision, CoT reasoning, and navigational inputs.

\noindent\textbf{Supervised Fine-Tuning with Formalized CoT Labels.} To activate the potential navigational reasoning ability of the LLM agent to adapt to the VLN task, we introduce the supervised-finetuning (SFT) paradigm in Stage 1 for conducting CoT training with our pre-constructed formalized CoT labels. 
Denote the navigation data sample at each timestep $t$ as $(P, \mathrm{CoT^{*}})$ (we omit the subscript $t$ for simplicity), where $P$ and $\mathrm{CoT^{*}}$ are the navigation prompt and CoT label, respectively. The training objective $\mathcal{L}_{\mathrm{SFT}}$ maximizes the likelihood of generating $\mathrm{CoT}^{*}$ given $P$ auto-regressively:
\begin{align}
\label{eq:sft loss}
\mathcal{L}_{\mathrm{SFT}}=-\mathbb{E}_{(P, \mathrm{CoT}^{*})\sim \mathcal{D}}\sum_{s=1}^{S}\mathrm{log}F_{LLM}(\mathrm{CoT}^{*}_{s}|P,\mathrm{CoT}^{*}_{<s}),
\end{align}
where $\mathcal{D}$ represents the navigation dataset. Denote the navigation action prediction training objective as $\mathcal{L}_{\mathrm{action}}$, the total training objective $\mathcal{L}_{\mathrm{Stage 1}}$ of Stage 1 is calculated as follows:
\begin{align}
\mathcal{L}_{\mathrm{Stage 1}}=\mathcal{L}_{\mathrm{action}}+\lambda\mathcal{L}_{\mathrm{SFT}},
\end{align}
where $\lambda$ represents the loss balance factor. We follow~\cite{zheng2024towards} to calculate the action prediction training objective $\mathcal{L}_{\mathrm{action}}$.

\noindent\textbf{Merits of formalized CoT labels.} 
Our design of formalized CoT labels has the following merits compared with free-form CoT labels like those collected in~\cite{zhou2024navgpt} (also see Figure~\ref{fig:overview}): Firstly, as the VLN task needs sequential decision making, generating formalized CoTs can significantly promote the reasoning speed compared to generating free-form ones. Secondly, free-form CoTs created by modern frontier models like GPT-4~\cite{zhou2024navgpt} may produce irrelevant and redundant reasoning for decision, while formalized CoT can produce concise and task-related reasoning. Thirdly, using formalized CoT labels for training can effectively simplify the training process as well as mitigate the hallucination compared to using free-form ones.

\begin{table*}[t]
\centering
\caption{Performance comparison results on R2R under the task-specific training setting. * denote our reimplementation results.  IL means the imitation learning setting. The best results for Cross-Modal Backbone and LLM-based Backbone are annotated in \textcolor{blue}{blue} and \textbf{bold} fonts, respectively.}
\renewcommand\arraystretch{1}
\resizebox{0.98\textwidth}{!}{
\setlength{\tabcolsep}{3.3mm}{
\begin{tabular}{l|cccc|cccc} \toprule
\multirow{2}{*}{Method}& \multicolumn{4}{c|}{Val Unseen} & \multicolumn{4}{c}{Test Unseen} \\
& TL & \textbf{NE}$\downarrow$ & \textbf{SR}$\uparrow$ & \textbf{SPL}$\uparrow$ & TL & \textbf{NE}$\downarrow$ & \textbf{SR}$\uparrow$ & \textbf{SPL}$\uparrow$  \\ \midrule
\multicolumn{9}{c}{{\it  \textbf{Cross-Modal Backbone}}:}\cr
\midrule
PREVALENT \cite{hao2020towards} & 10.19 & 4.71 & 58 & 53 & 10.51 & 5.30 & 54 & 51 \\
HOP \cite{qiao2022hop} &12.27 & 3.80 & 64 & 57 & 12.68 & 3.83 & 64 & 59 \\
HAMT \cite{Chen2021HistoryAM} & 11.46 & 2.29 & 66 & 61 & 12.27 & 3.93 & 65 & 60 \\
VLN-BERT \cite{hong2021vln} & 12.01 & 3.93 & 63 & 57  & 12.35 & 4.09 & 63 & 57 \\
DUET \cite{Chen2022ThinkGA} & 13.94 & 3.31 & 72 & 60 & 14.73 & 3.65 & 69 & 59 \\
Meta-Explore \cite{Hwang2023MetaExploreEH}& 13.09 & 3.22 & 72 & 62 & 14.25 & \textcolor{blue}{3.57} & 71 & \textcolor{blue}{61} \\
VLN-SIG \cite{li2023improving} & -& - & \textcolor{blue}{72} & \textcolor{blue}{62} & - & - & \textcolor{blue}{72} & 60 \\
VLN-PETL \cite{qiao2023vln} & 11.52 & 3.53 & 65 & 60 & 12.30 & 4.10 & 63 & 58 \\  
NavGPT2~\cite{zhou2024navgpt}& 13.25 & \textcolor{blue}{3.18} & 71 & 60 & - & - & - & - \\
\midrule
\multicolumn{9}{c}{{\it \textbf{LLM-based Backbone}}:}\cr
        \midrule
NavGPT~\cite{zhou2023navgpt}& 11.45 & 6.46 & 34 & 29 & - & - & - & - \\
DiscussNav~\cite{long2023discuss}& 9.69 & 5.32 & 43 & 40 & - & - & - & - \\
MapGPT~\cite{chen2024mapgpt}& - & 5.63 & 34 & 29 & - & - & - & - \\
NavCoT~\cite{lin2024navcot}& 9.95 & 6.26 & 40 & 37 & - & - & - & - \\
\hline
NaviLLM*~\cite{zheng2024towards} (IL)&9.99&6.04&46.90&43.78& 10.03 & 6.12 & 46 & 43\\
\rowcolor{cyan!15}
EvolveNav (IL, ours)&9.79&5.52&51.15&48.27& 9.94 &5.92&47&45\\
NaviLLM*~\cite{zheng2024towards}&13.43&3.27&70.11&60.25& 13.68 & 3.37 & 70 & 61 \\
\rowcolor{cyan!15}
\textbf{EvolveNav} (ours)&12.07&\textbf{3.15}&\textbf{71.17}&\textbf{63.48}&12.06&\textbf{3.22}&\textbf{71}&\textbf{63}\\
\bottomrule

\end{tabular}}
}

\label{table:r2r_sota}
\vspace{-0.2cm}
\end{table*}

\subsubsection{Stage 2: Self-Reflective Post-Training}
\label{Self-Reflective Post Training}
Although CoT supervised fine-tuning can explicitly guide the navigation agent to produce 
navigational reasoning for assisting the action decision, due to the uncertainty and complexity of the navigation task, using fixed labels may lead to overfitting to training CoT label distributions and therefore harm the generalization to unseen scenarios. Moreover, the inherent noise in the image captioning model~\cite{huang2023tag2text} for landmark detection may also limit the accuracy of formalized CoT labels collected in Stage 1. Therefore, after the Stage 1 training of Formalized CoT Supervised Fine-Tuning, we introduce Self-Reflective Post-Training to further encourage the navigation agent to learn correct reasoning patterns in a self-improving manner for improving generalization. 

\noindent\textbf{Self-Enriched CoT Label Enhancement.} 
To mitigate the overfitting to fixed CoT labels during training, we utilize the model's self-generated reasoning outputs as self-enriched CoT labels under the guidance of the model's action decision.
At timestep $t$, denote the model's reasoning output as $R_{t}$, the original formalized CoT label as $\mathrm{CoT}_{t}^{*}$, the model's action decision as $a_{t}$, and the ground-truth action  as $a_{t}^{*}$.
When the action decision $a_{t}$ generated by the navigation agent matches the ground-truth action $a_{t}^{*}$, we choose the agent's own reasoning output $R_{t}$ as the new CoT label. Such self-enriched CoT labels can effectively enhance the supervision diversity in a decision-oriented manner. 
Concretely, we obtain the updated CoT label $\tilde{\mathrm{CoT}}_{t}^{*}$ at timestep $t$ through the following rules:
\begin{align}
\tilde{\mathrm{CoT}}_{t}^{*}=
\begin{cases}
R_{t}, \quad \quad \text{if}  \quad a_{t}=a_{t}^{*}\\
\mathrm{CoT}_{t}^{*}, \quad \text{otherwise}
\end{cases}
\end{align}

\noindent\textbf{Self-Reflective Auxiliary Task.} To further make the model aware of correct and wrong reasoning, which can help the model better learn correct reasoning patterns, we additionally introduce a self-reflective auxiliary task, where we ask the model to discriminate which reasoning output from the given reasoning is right. Specifically, we collect positive and negative reasoning samples $R^{+}$ and $R^{-}$ during training for conducting the self-reflective auxiliary task. We utilize the above mentioned CoT label $\tilde{\mathrm{CoT}}_{t}^{*}$ as the positive reasoning sample $R^{+}$. To obtain the negative reasoning sample $R^{-}$, we randomly select the candidate observation (action) $O_{t,j}$ ($1<j<N$, $N$ is the number of navigable views) different from the ground-truth one. Then we extract the landmark and direction for $O_{t,j}$ to fill in the CoT label template (see Sec.~\ref{Formalized CoT Supervised Fine-Tuning}). As shown in Figure~\ref{fig:overview}, we construct the self-reflective task prompt $P_{\mathrm{sr}}$ as ``{\it Choose the correct one from the given two navigational reasoning outputs. Output 1: [$R^{1}$]. Output 2: [$R^{2}$]. Selection:} '', where we randomly insert the positive reasoning sample $R^{+}$ and negative reasoning sample $R^{-}$ to the positions of $R^{1}$ and $R^{2}$. We collect the ground-truth output  $R_{\mathrm{sr}}^{*}$ with the form like ``Output 2.''. We also utilize the auto-regressive training objective like Eq.~\ref{eq:sft loss} to calculate the loss $\mathcal{L}_{\mathrm{sr}}$ based on ($P_{\mathrm{sr}}$, $R_{\mathrm{sr}}^{*}$) pairs for the self-reflective auxiliary task. 

With the self-reflective loss $\mathcal{L}_{\mathrm{sr}}$, the total training objective $\mathcal{L}_{\mathrm{Stage 2}}$ for Stage 2 is obtained by:
\begin{align}
\mathcal{L}_{\mathrm{Stage 2}}=\mathcal{L}_{\mathrm{action}}+\lambda_{1}\mathcal{L}_{\mathrm{SFT}}+\lambda_{2}\mathcal{L}_{\mathrm{sr}},
\end{align}
where both $\lambda_{1}$ and $\lambda_{2}$ are the loss coefficients. Through the self-enriched CoT label enhancement and self-reflective auxiliary task, the navigation agent learns to generate correct embodied reasoning in a self-refining manner to enable adaptable reasoning in different scenarios.

\subsubsection{Training and Inference}

\noindent\textbf{Two-stage training. }
When conducting training in the proposed self-improving embodied reasoning framework, we train the model in Stage 1 to converge, and use it for Stage 2 training. This two-stage training design  is to mitigate the negative impact of noisy CoT outputs during Stage 1 under a  non-converged situation and ensure the training stability during Stage 2. Specifically, in the early training iterations during Stage 1 of Formalized CoT Supervised Fine-Tuning, the agent is prone to generate CoT reasoning with noisy formats and information (e.g., its output may contain repeatedly notions of “$<$s$>$” or the output may be very long) while accompanied action decisions may be correct occasionally. Based on the rule of our self-enriched label enhancement strategy during Stage 2, such noisy outputs will be introduced as CoT labels to guide the CoT training and may cause training instability. By firstly training the model in Stage 1 to converge, such issues can be effectively alleviated. This two-stage training manner also simplifies the method design and implementation to promote its practicality during real deployment.

\noindent\textbf{Inference.}
During inference, the model generates CoT based on the given prompt with the form of “-Action Decision: $<$cls$>$. -Navigational Reasoning:”. The encoding of the $<$cls$>$ token is extracted to generate the action prediction probability $a_{t}$ through the action prediction head $F_{\mathrm{action}}$ (see Sec.~\ref{Model Architecture}). 


	
\begin{table}
\caption{Performance comparison results on CVDN under the task-specific training setting. We utilize the Goal Progress (GP) (m) as the evaluation metric. * denote our reimplementation results.  The best results for Cross-Modal Backbone and LLM-based Backbone are annotated in \textcolor{blue}{blue} and \textbf{bold} fonts, respectively.}
\centering\resizebox{\linewidth}{!}{
    {
\renewcommand\arraystretch{1}
\setlength{\tabcolsep}{5mm}{
\begin{tabular}{l|cc} \hline
Method& Val-Unseen & Test \\ \midrule
\multicolumn{3}{c}{{\it \textbf{Cross-Modal Backbone}}:}\cr
        \midrule
Seq2Seq \cite{thomason2019vision} & 2.10 & 2.35 \\
PREVALENT \cite{hao2020towards} & 3.15 & 2.44 \\
HOP \cite{qiao2022hop} & 4.41 & 3.31 \\
MT-RCM \cite{wang2020environment} & 4.36 & - \\
MT-RCM+Env \cite{wang2020environment} & 4.65 & 3.91 \\
HAMT \cite{Chen2021HistoryAM} & 5.13 & 5.58 \\
VLN-SIG \cite{li2023improving} & 5.52 & 5.83 \\
VLN-PETL \cite{qiao2023vln} & \textcolor{blue}{5.69} & \textcolor{blue}{6.13} \\
\midrule
\multicolumn{3}{c}{{\it \textbf{LLM-based Backbone}}:}\cr
        \midrule
NaviLLM*~\cite{zheng2024towards}&5.53& 6.80\\
\rowcolor{cyan!15}

\textbf{EvolveNav} (ours) &\textbf{6.21}&\textbf{7.07}\\
\bottomrule
\end{tabular}}}}

\label{table:cvdn_sota}
\end{table}

\section{Experiment}
\subsection{Experimental Setup}
\label{Experimental Setup}

\subsubsection{Datasets}
We test EvolveNav on four popular VLN benchmarks, i.e., R2R~\cite{anderson2018vision}, CVDN~\cite{thomason2019vision}, REVERIE~\cite{qi2020reverie}, and SOON~\cite{zhu2021soon}. Each benchmark handles distinct
challenges posed by VLN. 
\textbf{R2R} is built on 90 real-world indoor simulation environments containing 7,189 trajectories, each corresponding to three fine-grained instructions. \textbf{CVDN} contains 2,050 human-human navigation dialogs and over 7k trajectories in 83 MatterPort houses. \textbf{REVERIE} replaces the
fine-grained instructions in R2R with high-level instructions. \textbf{SOON} constructs thoroughly described instructions to further highlight visual-semantic alignment.

To verify the effectiveness of EvolveNav, we adopt two representative training setting, task-specific training and cross-task training. Task-specific training trains the model on single benchmark like most previous works~\cite{Chen2021HistoryAM,Chen2022ThinkGA,lin2024navcot}, while cross-task training realizes a generalist navigation model like NaviLLM~\cite{zheng2024towards} by training the model using multiple datasets.


\subsubsection{Evaluation Metrics}
We utilize the following standard metrics for evaluation: 1) Trajectory Length (\textbf{TL}): the average length of the agent's navigated path, 2) Navigation Error (\textbf{NE}): the average distance between the agent's destination and the goal viewpoint, 3) Success Rate (\textbf{SR}): the ratio of success, where the agent stops within three meters of the target point, 4) Success rate weighted by Path Length (\textbf{SPL})~\cite{anderson2018vision}: success rate normalized by the ratio between the length of the shortest path and the predicted path, 5) Oracle Success Rate (\textbf{OSR}): the ratio of containing a viewpoint along the path where the target position is visible, 6) Goal Progress (\textbf{GP}), the progress in meters towards the goal. 

\subsubsection{Implementation Details}
We utilize NaviLLM~\cite{zheng2024towards} as our baseline model. To simplify the implementation, we do not introduce the pretraining phase in~\cite{zheng2024towards} for both our EvolveNav and the baseline (denoted as NaviLLM* in Table~\ref{table:r2r_sota}-~\ref{table:multitask_sota}).
In our EvolveNav,
We fine-tune the LLM with full-parameter and LoRA settings for Stage 1 and 2 training, respectively.
The training for Stage 1 is conducted on 8 Nvidia A100 GPUs and the training for Stage 2 is performed on 4 Nvidia A100 GPUs. Empirically, we set the loss coefficients $\lambda$, $\lambda_{1}$, and $\lambda_{2}$ as 1, 1, and 0.2, respectively. During Stage 1 training, we introduce the CoT supervised finetuning loss $\mathcal{L}_{\mathrm{SFT}}$ under a probability of 0.5 to mitigate the overfitting to the pre-constructed CoT labels.
The maximum numbers of training steps for Stage 1 and 2 are set as 60000 and 9000 steps, respectively. Training for Stage 1 with full parameter lasts for $\sim$1.5 days with $\sim$73G GPU memory, and training for Stage 2 with LoRA lasts for $\sim$1 day with $\sim$30G GPU memory.
The hyperparameters such as the learning rate, optimizer, and the batch size are kept the same as~\cite{zheng2024towards}.

\begin{table}
\caption{Performance comparison results on SOON under the task-specific training setting. * denote our reimplementation results.  The best results for Cross-Modal Backbone and LLM-based Backbone are annotated in \textcolor{blue}{blue} and \textbf{bold} fonts, respectively.}
\centering

\resizebox{1.0\linewidth}{!}{{
\renewcommand\arraystretch{1}
\setlength{\tabcolsep}{3mm}{
\begin{tabular}{l|ccc} \toprule

\multirow{2}{*}{Method}&\multicolumn{3}{c}{Val-Unseen} \\ 
 & \textbf{OSR}$\uparrow$ & \textbf{SR}$\uparrow$ & \textbf{SPL}$\uparrow$ \\ \midrule
 \multicolumn{4}{c}{{\it \textbf{Cross-Modal Backbone}}:}\cr\midrule
GBE~\cite{zhu2021soon} & 28.54 & 19.52 & 13.34   \\ 
DUET~\cite{Chen2022ThinkGA}  & 50.91 &  36.28 & 22.58 \\
AZHP~\cite{gao2023adaptive} &\textcolor{blue}{56.19} & \textcolor{blue}{40.71} & \textcolor{blue}{26.58} \\ 
\midrule
\multicolumn{4}{c}{{\it \textbf{LLM-based Backbone}}:}\cr\midrule
NaviLLM*~\cite{zheng2024towards}  & 43.60 & 30.34 & 23.70  \\
\rowcolor{cyan!15}\textbf{EvolveNav}~(Ours)  & \textbf{49.56} & \textbf{33.40} & \textbf{24.92}  \\

\bottomrule
\end{tabular}}
}}
\label{table:soon_sota}
\end{table}

\begin{table}[t]
\caption{Performance comparison results on REVERIE under the task-specific training setting. * denote our reimplementation results.  The best results for Cross-Modal Backbone and LLM-based Backbone are annotated in \textcolor{blue}{blue} and \textbf{bold} fonts, respectively.}
\centering
\renewcommand\arraystretch{1}
\resizebox{\linewidth}{!}{
\setlength{\tabcolsep}{3.mm}{
\begin{tabular}{l|ccc}
\toprule
\multirow{2}{*}{Method}& \multicolumn{3}{c}{Val Unseen} \\
& \textbf{OSR}$\uparrow$ & \textbf{SR}$\uparrow$ & \textbf{SPL}$\uparrow$ \\
\midrule
\multicolumn{4}{c}{{\it \textbf{Cross-Modal Backbone}}:}\cr\midrule
Seq2Seq~\cite{anderson2018vision}  & 8.07 & 4.20 & 2.84  \\
HOP~\cite{qiao2022hop} & 36.24 & 31.78 & 26.11 \\
HAMT~\cite{Chen2021HistoryAM}  & 36.84 & 32.95 & 30.20  \\
VLN-BERT~\cite{hong2021vln}  & 35.02 & 30.67 & 24.90  \\
DUET~\cite{Chen2022ThinkGA}  & 51.07 & 46.98 & 33.73  \\
AZHP~\cite{gao2023adaptive} & \textcolor{blue}{53.65} & \textcolor{blue}{48.31} & \textcolor{blue}{36.63}  \\
VLN-PETL~\cite{qiao2023vln}  & 37.03 & 31.81 & 27.67  \\
\midrule
\multicolumn{4}{c}{{\it \textbf{LLM-based Backbone}}:}\cr\midrule
NaviLLM*~\cite{zheng2024towards} & \textbf{42.68} & 32.55 & 25.82  \\
\rowcolor{cyan!15}\textbf{EvolveNav}~(Ours)   & \underline{42.40} & \textbf{33.60} & \textbf{28.16}  \\ 
\bottomrule
\end{tabular}}
}

\label{table:reverie_sota}
\end{table}


\subsection{Comparison with Existing Methods}
\noindent\textbf{Task-specific training.}
Table~\ref{table:r2r_sota},~\ref{table:cvdn_sota},~\ref{table:soon_sota}, and~\ref{table:reverie_sota} present the task-specific training results on R2R~\cite{anderson2018vision}, CVDN~\cite{thomason2019vision}, SOON~\cite{zhu2021soon}, and REVERIE~\cite{qi2020reverie}, respectively. The results exhibit consistent superiority of EvolveNav over the compared approaches on various VLN benchmarks, demonstrating the effectiveness and excellent generalization ability of the proposed self-improving embodied reasoning paradigm. For example, for the results on R2R in Table~\ref{table:r2r_sota}, the performance gain in SPL of EvolveNav on Val Unseen is $\sim$4.5\% and $\sim$3.2\% under the imitation learning (IL) and dagger~\cite{Chen2022ThinkGA} training settings compared to the baseline model NaviLLM~\cite{zheng2024towards}, respectively. For the results on SOON in Table~\cite{zhu2021soon}, the performance improvements of EvolveNav over the baseline model NaviLLM on OSR, SR, and SPL are $\sim$5.9\%, $\sim$3.1\%, and $\sim$1.2\%, respectively. Note that we do not consider the comparison with models augmented by new environments (e.g., ScaleVLN~\cite{wang2023scaling}) for fairness.

\begin{table*}
\centering
\caption{Performance comparison results under the cross-task training setting. * denote our reimplementation results.  The best results for task-specific training (``Separate Model for Each Task'') and cross-task training (``Unified Model for All Tasks'') are annotated in \textcolor{blue}{blue} and \textbf{bold} fonts, respectively.}
\renewcommand\arraystretch{1}
\resizebox{0.98\textwidth}{!}{
\setlength{\tabcolsep}{3.3mm}{
\begin{tabular}{l|ccc|ccc|ccc|c} \toprule
\multirow{2}{*}{Method}& \multicolumn{3}{c|}{REVERIE} & \multicolumn{3}{c|}{SOON}&\multicolumn{3}{c|}{R2R} &CVDN\\
&  \textbf{SR}$\uparrow$ &\textbf{OSR}$\uparrow$ & \textbf{SPL}$\uparrow$ &\textbf{SR}$\uparrow$ &\textbf{OSR}$\uparrow$ & \textbf{SPL}$\uparrow$&\textbf{SR}$\uparrow$ &\textbf{OSR}$\uparrow$ & \textbf{SPL}$\uparrow$&\textbf{GP}$\uparrow$ \\ \midrule
\multicolumn{11}{c}{{\it \textbf{Separate Model for Each Task}}:}\cr\midrule
DUET~\cite{Chen2022ThinkGA}&46.98&51.07&33.73&36.28&50.91&22.58&\textcolor{blue}{72}&-&60&-\\
AZHP~\cite{gao2023adaptive}&\textcolor{blue}{48.31}&\textcolor{blue}{53.65}&\textcolor{blue}{36.63}&\textcolor{blue}{40.71}&\textcolor{blue}{56.19}&\textcolor{blue}{26.58}&-&-&-&-\\
VLN-PETL~\cite{qiao2023vln}&31.81&37.03&27.67&-&-&-&65&-&60&5.69\\
NaviLLM*~\cite{zheng2024towards}&32.55&42.68&25.82&30.34&43.60&23.70&70.11&\textcolor{blue}{79.00}&60.25&5.53\\
\textbf{EvolveNav}~(Ours)&33.60&42.40&28.16&33.40&49.56&24.92&71.17&78.95&\textcolor{blue}{63.48}&\textcolor{blue}{6.21}\\
\midrule
\multicolumn{11}{c}{{\it \textbf{Unified Model for All Tasks}}:}\cr\midrule
NaviLLM~\cite{zheng2024towards}&44.56&53.74&36.63&35.44&-&28.09&67&-&58&5.91\\
NaviLLM*~\cite{zheng2024towards}&43.81&54.26&35.61&34.82&55.01&26.19&\textbf{68.37}&76.96&\textbf{59.09}&5.43\\
\rowcolor{cyan!15}\textbf{EvolveNav}~(Ours)&\textbf{45.97}&\textbf{57.95}&\textbf{38.58}&\textbf{37.00}&\textbf{58.20}&\textbf{28.18}&\underline{68.07}&\textbf{81.68}&\underline{58.30}&\textbf{6.35}\\
\bottomrule

\end{tabular}}
}

\label{table:multitask_sota}
\end{table*}

\noindent\textbf{Cross-task training.}
Table~\ref{table:multitask_sota} shows the cross-task training results on R2R~\cite{anderson2018vision}, CVDN~\cite{thomason2019vision}, SOON~\cite{zhu2021soon}, and REVERIE~\cite{qi2020reverie}. From Table~\ref{table:multitask_sota}, we can observe that EvolveNav surpasses the baseline approach NaviLLM~\cite{zheng2024towards} in most metrics on different benchmarks. These results reveal that our self-improving embodied reasoning framework is also effective for training the generalist navigation model, which is more practical and flexible in real-world navigation scenarios. Both the task-specific and cross-task training results on various VLN benchmarks sufficiently demonstrate that the proposed self-improving embodied reasoning framework fulfills adaptable and generalizable navigational reasoning under different tasks and scenarios.


\subsection{Ablation Study}

\noindent\textbf{Effect of different method components.} 
Table~\ref{tab:ablation} presents the ablation study results on Val Unseen set on R2R, where we can find the effectiveness and reasonability of different method components in our EvolveNav. From Table~\ref{tab:ablation}, we can observe that through the Stage 1 training of formalized CoT supervised fine-tuning (SFT) (``1''), the model's navigational reasoning ability can be significantly enhanced to improve the navigation performance. In Stage 2 training, the introduction of self-enriched CoT labels (``2'') and the self-reflective auxiliary task (``3'') can further bring performance gain respectively in both SR and SPL metrics compared to pure CoT SFT in Stage 1. Our full model (``Full Model'') achieves the best results in SR and SPL compared with ``2'' and ``3'', demonstrating that the combination of our self-enriched CoT label enhancement strategy and self-reflective auxiliary task is non-trivial. Especially, rather than pure self-reflective auxiliary task (``3'') that asks the agent to learn to distinguish fixed correct and wrong reasoning patterns like conventional auxiliary task design, the combination of our self-enriched CoT label enhancement strategy and self-reflective auxiliary task (``Full Model'') can encourage the agent to learn diverse correct reasoning and therefore increase its generalization ability to unseen scenarios.

\begin{table}
    \caption{Ablation study of method components on Val Unseen set on R2R. We adopt the imitation learning (IL) setting for evaluation. ``CoT SFT'' represents the Stage 1 training of Formalized CoT Supervised-Finetuning. ``Self-Enriched CoT SFT'' denote the Self-Enriched CoT Label Enhancement strategy in Stage 2 training of Self-Reflective Post-Training.
    }
\fontsize{28}{28}\selectfont
    \centering
    \resizebox{\linewidth}{!}{
    {\renewcommand{\arraystretch}{1.2}
    \begin{tabular}{c|ccc|ccc}
    \specialrule{.1em}{.05em}{.05em}
   
   \multirow{2}{*}{Method}&Stage 1&\multicolumn{2}{c|}{Stage 2}&\multicolumn{3}{c}{Val Unseen}\\
   &\makecell{CoT \\SFT}&\makecell{Self-Enriched\\ CoT SFT}&\makecell{Self-Reflective \\Auxiliary Task} &SR$\uparrow$&OSR$\uparrow$&SPL$\uparrow$\\

			
        \hline
        Baseline&-&-&-&46.90&54.63&43.78\\
        1 &\checkmark&&&49.62&\textbf{59.44}&46.26\\
        2 &\checkmark&\checkmark&&50.47&57.48&47.98\\
        3&\checkmark&&\checkmark&50.51&57.74&47.86\\
        Full Model&\checkmark&\checkmark&\checkmark&\textbf{51.15}&\underline{59.18}&\textbf{48.27}\\
 \specialrule{.1em}{.05em}{.05em}
    \end{tabular}
}}
\label{tab:ablation}
\end{table}

\begin{figure*}[t]
\begin{centering}
\includegraphics[width=0.98\linewidth]{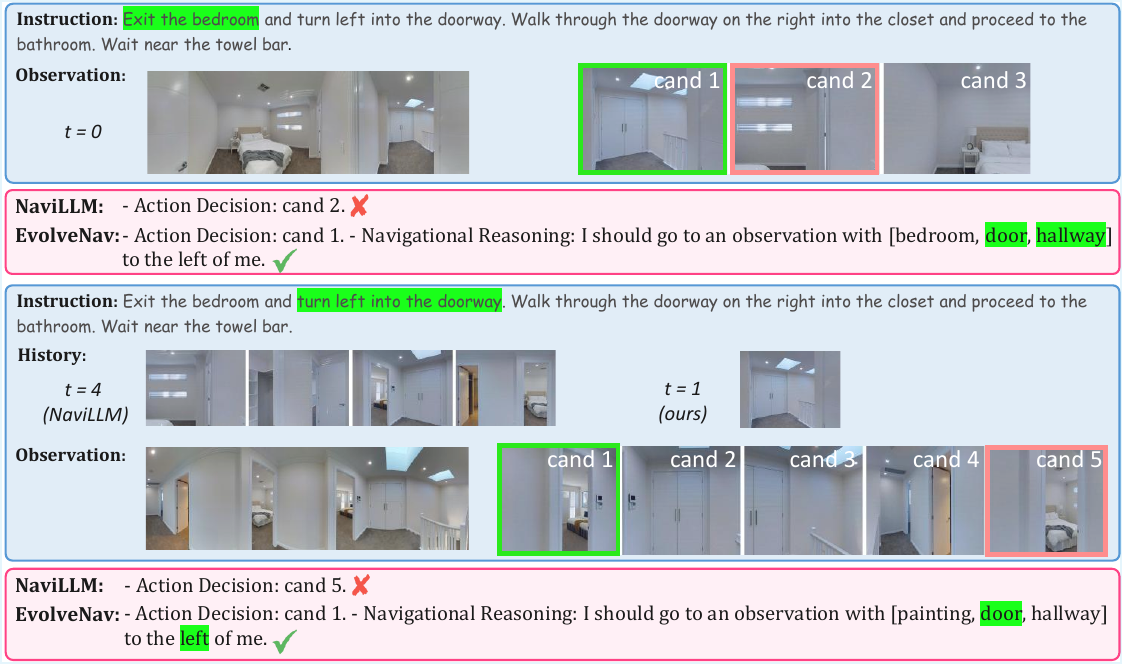}
\par\end{centering}
\caption{Action decision visualization of NaviLLM~\cite{zheng2024towards} and our EvolveNav. We only extract two steps and display local candidate space for simplicity. Observations selected by EvolveNav (also are the ground-truth actions) and NaviLLM are annotated by green boxes and red boxes, respectively.}
\vspace{-0.4cm}
\label{fig:comparison-with-navillm_cropped}
\end{figure*}

\noindent\textbf{Effect of constructed CoT labels.} Table~\ref{tab:ablation cot} compares the navigation performance under different kinds of CoT labels, where we can find the effectiveness of our introduced CoT labels by predicting landmark and direction information in a formalized way. To realize ``Free-form CoT'', we introduce the free-form CoT labels collected in NavGPT-2~\cite{zhou2024navgpt} to train the navigation agent. To realize ``Direction \& Landmark$^{\dag}$'', we obtain the best matched landmark in the instruction to each ground-truth observation through the CLIP model~\cite{radford2021learning}. 

From Table~\ref{tab:ablation cot}, we can draw into the following conclusions: 1) The comparison between ``Free-form CoT'' and ``Direction \& Landmark (ours)'' shows that our 
formalized CoT labels can effectively reduce redundant reasoning information to improve the navigational reasoning and decision accuracy. Moreover, the inference time to generate CoT at one timestep of ``Free-form CoT'' is $\sim$7.8s compared to $\sim$2.5s of ``Direction \& Landmark (ours)'', demonstrating that our method can significantly promote the reasoning speed ($\sim$$\times$3 improvement), which is crucial for sequential decision making task like navigation. 2) The comparison among ``Only Direction'', ``Only Landmark'', and ``Direction \& Landmark (ours)'' reveal that both landmark and direction information are important to guide navigation decisions, demonstrating the reasonability of our constructed CoT labels. 
3) The superiority of ``Direction \& Landmark (ours)'' over ``Direction \& Landmark$^{\dag}$'' demonstrates that our introduced CoT labels, which contain diverse landmarks, can potentially encourage the navigation agent to learn cross-modal alignment knowledge to accurately align the visual observation to the navigation instruction.    
\begin{figure*}
\begin{centering}
\includegraphics[width=0.98\linewidth]{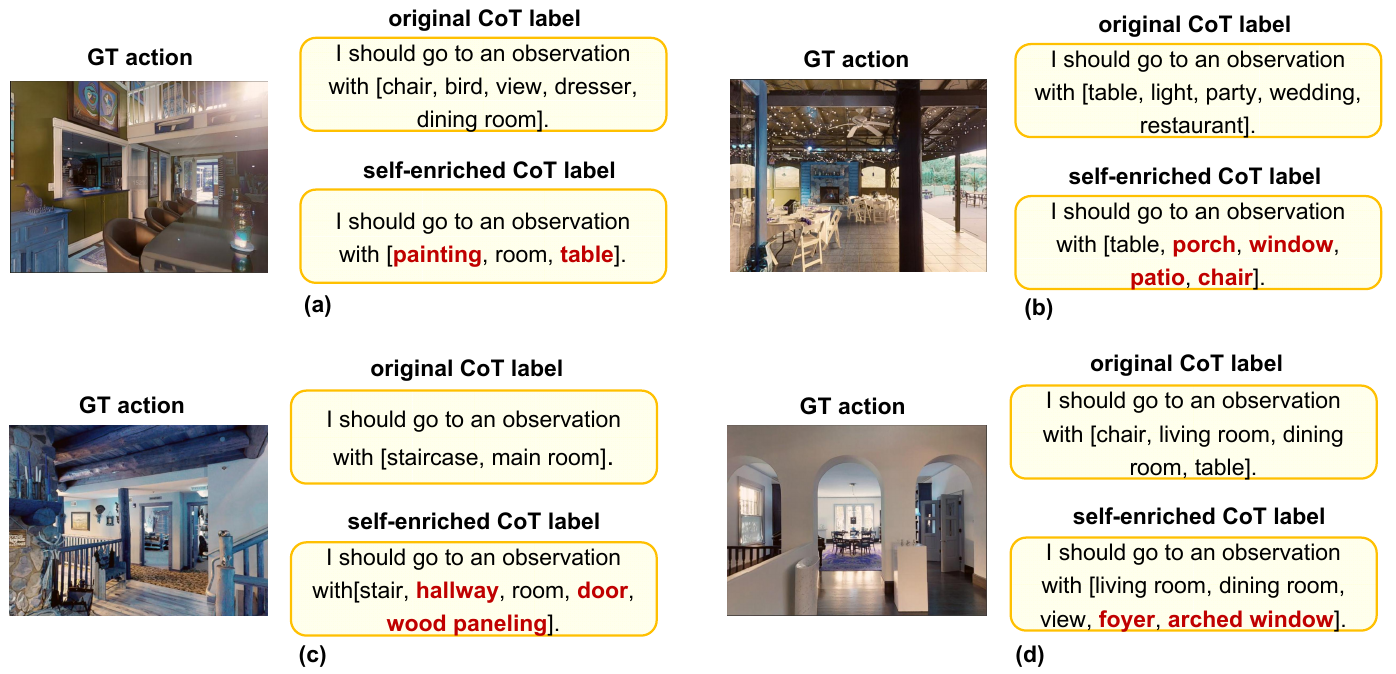}
\par\end{centering}
\caption{Visualization comparison between self-enriched chain-of-thought (CoT) labels and originally built CoT labels. Newly introduced landmarks in the self-enriched CoT label are highlighted in \textcolor{red}{red} fonts. GT action denotes the ground-truth action (observation). We omit the direction information in the CoT labels.}
\label{fig:visualization-supp-label}
\end{figure*}

\begin{table}
    \caption{Ablation study of CoT labels on Val Unseen set on R2R. $^{\dag}$ denotes using the landmark mentioned in the instruction.}
    \small
    \centering
    \resizebox{\linewidth}{!}{
    {\renewcommand{\arraystretch}{1.0}
    \begin{tabular}{l|ccc}
    \specialrule{.1em}{.05em}{.05em}
			Method&SR$\uparrow$&OSR$\uparrow$&SPL$\uparrow$\cr

			
        \hline
        Baseline&70.11&79.00&60.25\\
        \hline
        Free-form CoT&69.39&\textbf{80.44}&60.95\\
        Only Direction&67.05&76.49&57.32\\
        Only Landmark&68.32&79.85&60.53\\
        Direction \& Landmark$^{\dag}$&69.77&79.25&60.77\\
        Direction \& Landmark (ours)&\textbf{71.26}&\underline{80.23}&\textbf{62.05}\\
 \specialrule{.1em}{.05em}{.05em}
    \end{tabular}
}}
\label{tab:ablation cot}
\end{table}

\subsection{Visualization}
In this subsection, we present various kinds of visualization results, including visualization of action decision, self-enriched CoT label, loss \& performance variation, and landmark extraction, to comprehensively and deeply analyze the advantage of the proposed self-improving embodied reasoning framework.

\noindent\textbf{Action Decision Visualization.}
Fig.~\ref{fig:comparison-with-navillm_cropped} gives the action decision visualization comparison between NaviLLM~\cite{zheng2024towards} and our EvolveNav, where we can find that EvolveNav generates reasonable navigational reasoning about landmarks and directions to guide correct action decision making.
For example, when $t=0$, from the observations, EvolveNav infers that an observation with \textit{door} and \textit{hallway} represents the exit from the bedroom while NaviLLM mistakenly selects an action that remains in the bedroom. 
Another example is in the hallway ($t=4$ for NaviLLM and $t=1$ for EvolveNav), EvolveNav generates reasoning consistent with the decision of entering the correct \textit{doorway} on the \textit{left}. However, NaviLLM chooses the wrong doorway to go back. 
These results highlight the effectiveness of our approach in improving navigational reasoning for accurate instruction understanding and action prediction. 

\begin{figure*}	\setlength{\abovecaptionskip}{3pt}
	\setlength{\belowcaptionskip}{3pt}
	\centering
	\renewcommand{\figurename}{Figure}
	\subfloat[Loss variation]{
		\includegraphics[width=8.8cm,height=6.5cm]{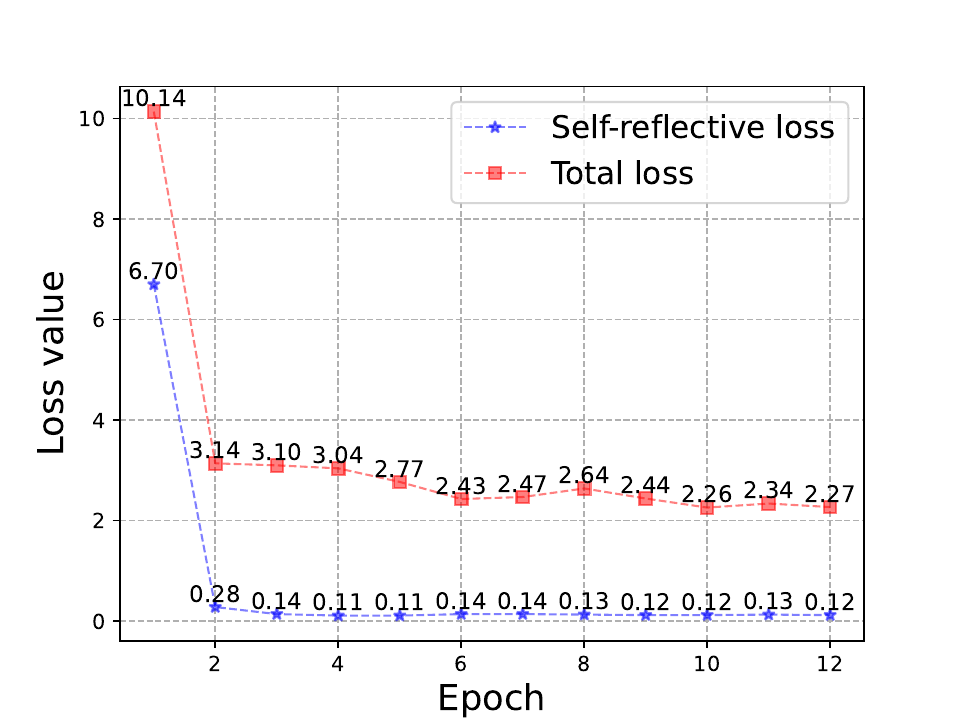}}
	\hspace{-0.01in}
	\subfloat[Navigation Error (NE) variation]{
		\includegraphics[width=8.8cm,height=6.5cm]{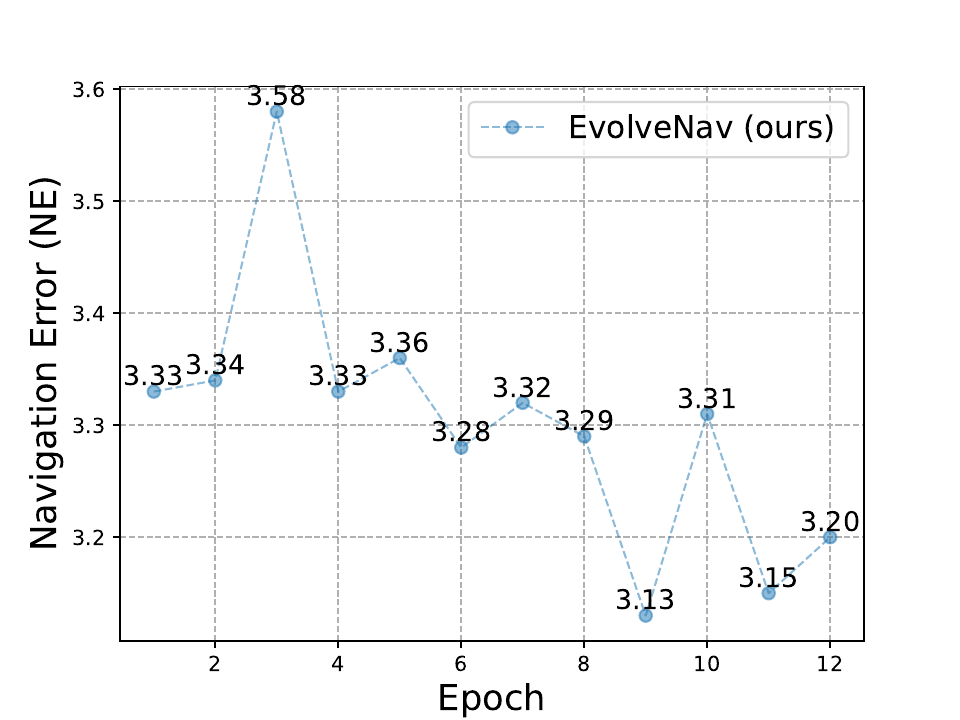}}
	
	\caption{Loss and performance variation during Stage 2: Self-Reflective Post-Training. Low navigation error (NE) value indicates better results.}
	\label{fig:learning curve} 
\end{figure*}

\begin{figure*}[t]
\begin{centering}
\includegraphics[width=1.0\linewidth]{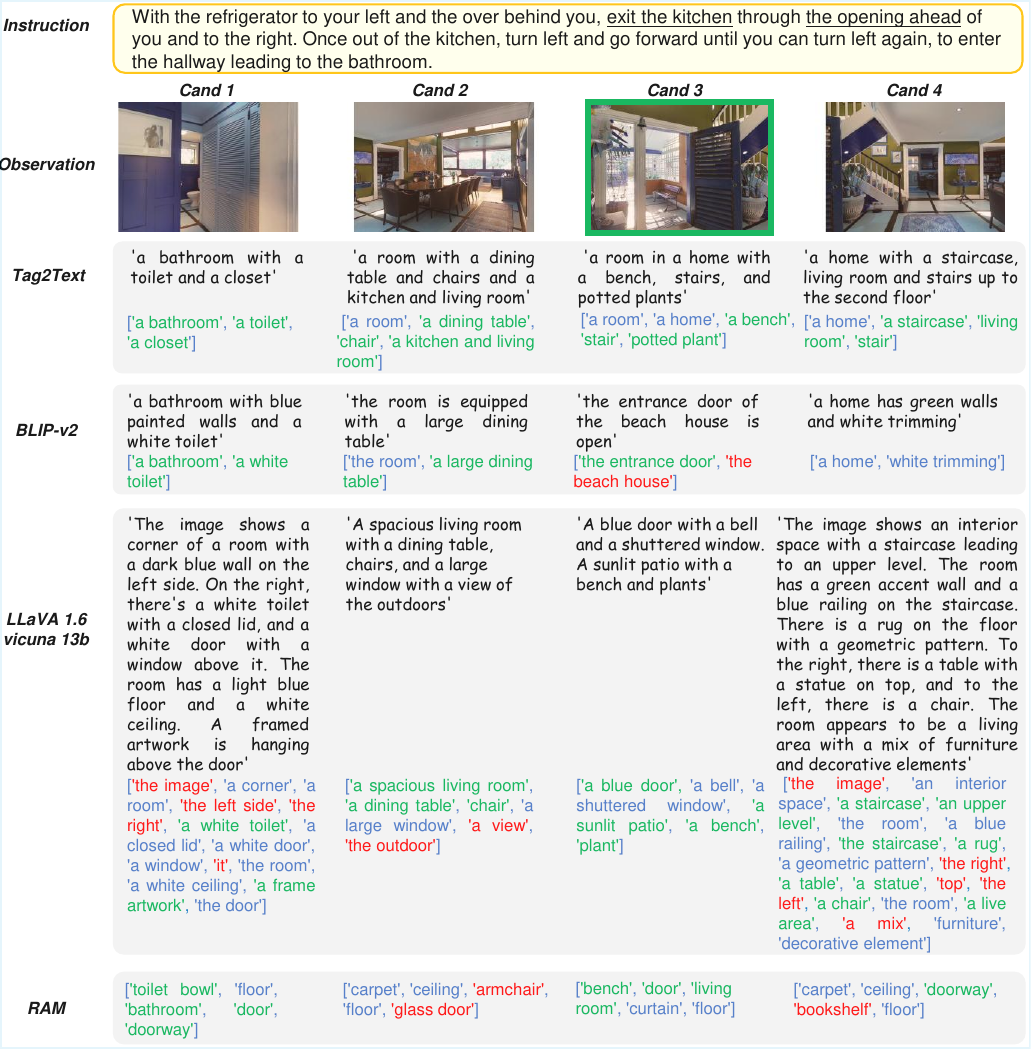}
\par\end{centering}
\caption{Landmark extraction visualization of different methods. We use \textcolor{green}{green}, \textcolor{red}{red}, and \textcolor{blue}{blue} colors to distinguish informative, false, and uninformative landmarks, respectively.}
\label{fig:visualization-supp-landmark}
\end{figure*}

\noindent\textbf{Self-Enriched CoT label visualization.}
In Fig.~\ref{fig:visualization-supp-label}, we present some visualization comparison examples of original CoT labels and self-enriched CoT labels. From Fig.~\ref{fig:visualization-supp-label}, we can observe that our self-enriched CoT label enhancement strategy effectively increases the supervision diversity. For example, in Fig.~\ref{fig:visualization-supp-label}(a) and (b), the self-enriched CoT labels from the model's own reasoning outputs capture the landmark {\it painting} and {\it chair} which are not contained in the original CoT label, respectively. From Fig.~\ref{fig:visualization-supp-label}(c) and (d), we can find that the LLM-based navigation agent can also recognize the attribute of the landmark, e.g., it recognizes {\it wood paneling} and {\it arched window}, which indicate the material and shape of the landmark, respectively. Such CoT labels can help navigation agent learn to follow more fine-grained instructions. Benefiting from our self-enriched CoT label, the LLM-based navigation agent can reduce the overfitting to the original CoT label distributions and learn more diverse cross-modal alignment knowledge, and therefore promote the generalization to unseen scenarios.

\noindent\textbf{Loss \& performance variation during Stage 2 training.}
Fig.~\ref{fig:learning curve} shows the loss and performance curves during Self-Reflective Post-Training (Stage 2). In Fig.~\ref{fig:learning curve}, we can find that our self-reflective loss $\mathcal{L}_{\mathrm{sr}}$, the total training loss $\mathcal{L}_{\mathrm{Stage 2}}$, and the navigation error (NE) have similar variation trends. Especially, both two loss curves and the NE curves achieve the lowest value around epoch 9. These results show the effectiveness of our constructed self-reflective auxiliary task during Self-Reflective Post-Training in improving the navigational reasoning and decision accuracy of the LLM-based navigation agent.

\noindent\textbf{Landmark extraction visualization.}
We compare different methods for constructing the formalized CoT labels, to verify the reasonability of our landmark extraction strategy by combining image captioning model~\cite{huang2023tag2text} with the NLP tool Spacy~\cite{spacy}. 
Fig.~\ref{fig:visualization-supp-landmark} presents the landmark extraction visualization comparison of 
three image captioning models, 
Tag2Text~\cite{huang2023tag2text}, BLIP-v2~\cite{li2023blip2}, LLaVA 1.6 vicuna 13b~\cite{liu2024llavanext},  and an open-vocabulary object recognition model RAM~\cite{zhang2024ram}.
Concretely, we provide four candidates in a navigational step for these methods and use Spacy~\cite{spacy} to extract the landmarks in the output caption (except for RAM of directly obtaining tagging). 

From Fig.~\ref{fig:visualization-supp-landmark}, we can observe that Tag2Text can capture more informative landmarks while having less redundancy and illusion. For example, for ``Cand 2'', Tag2Text correctly detects \textit{a kitchen and living room}, \textit{a dining table}, and \textit{chair},  while BLIP-v2 only detects \textit{a large dining table}. Although LLaVA 1.6 13b generates abundant captions, it brings noisy landmarks like \textit{the outdoor} after noun phrases extraction. RAM also generates meaningless and non-existent landmarks, like \textit{floor} and \textit{armchair}.
These results show that our combination of Tag2text model~\cite{huang2023tag2text} and NLP tool for landmark extraction can effectively retain informative landmarks while reducing redundancy and illusion for constructing CoT labels, which can enable the agent to better learn cross-modal alignment between observations and instructions.

\section{Conclusion}
In this work, we propose \textbf{EvolveNav}, a novel self-improving embodied reasoning framework to fulfill generalizable and adaptable reasoning for enhancing LLM-based vision-and-language navigation. Through introducing the formalized CoT supervised fine-tuning and self-reflective post-training in the proposed framework, the agent's navigational reasoning ability can be effectively enhanced while mitigating the overfitting to the training reasoning label distributions simultaneously to improve generalization. 
Experimental results on multiple VLN benchmarks under diverse training settings reveal the promising capability of our method in boosting the reasoning ability and decision accuracy for LLM-based navigation agents.
We believe that our EvolveNav can provide meaningful references for designing self-improving embodied reasoning paradigms to benefit future LLM-assisted Embodied AI research.

\bibliography{egbib}

\bibliographystyle{IEEEtran}
\bibdata{IEEEabrv,egbib}	

\begin{IEEEbiography}[{\includegraphics[width=1in,height=1.25in,clip,keepaspectratio]{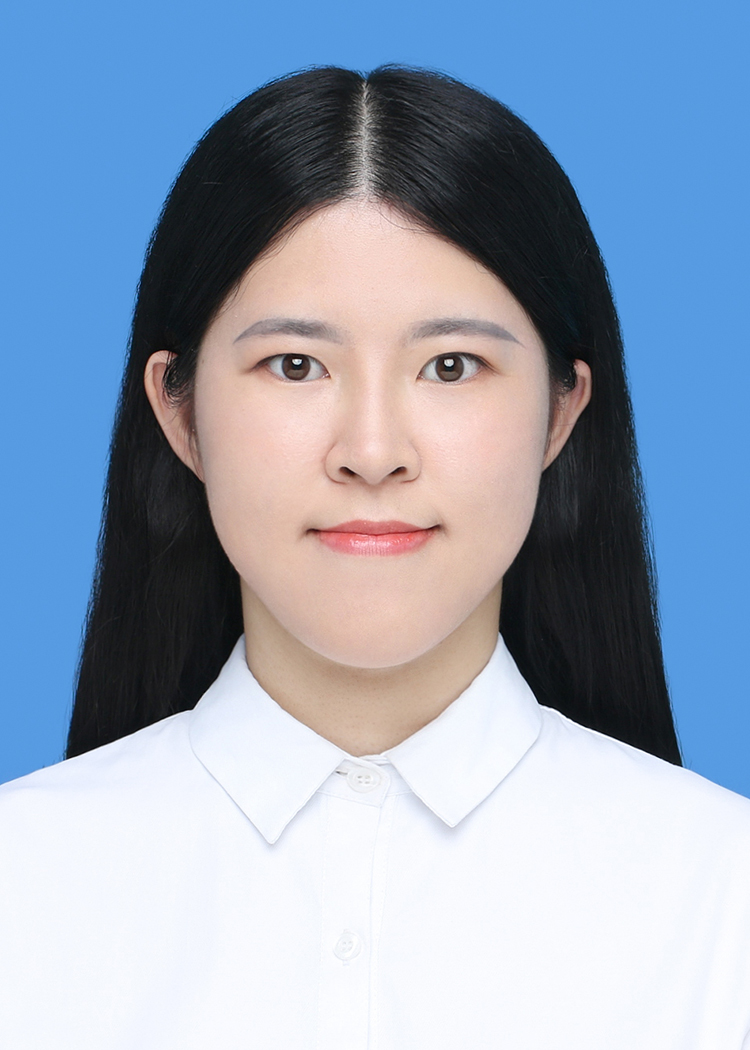}}]
{Bingqian Lin} is currently a postdoc researcher at Shanghai Jiao Tong University, advised by Prof. Cewu Lu. She received her PhD degree from Sun Yat-sen University in 2024, advised by Prof. Xiaodan Liang and Prof. Liang Lin. She received the B.E. and the M.E.
degree in Computer Science from University of
Electronic Science and Technology of China and
Xiamen University, in 2016 and 2019, respectively.
 Her research interests include vision-and-language understanding and embodied AI.
\end{IEEEbiography}

\begin{IEEEbiography}[{\includegraphics[width=1in,height=1.25in,clip,keepaspectratio]{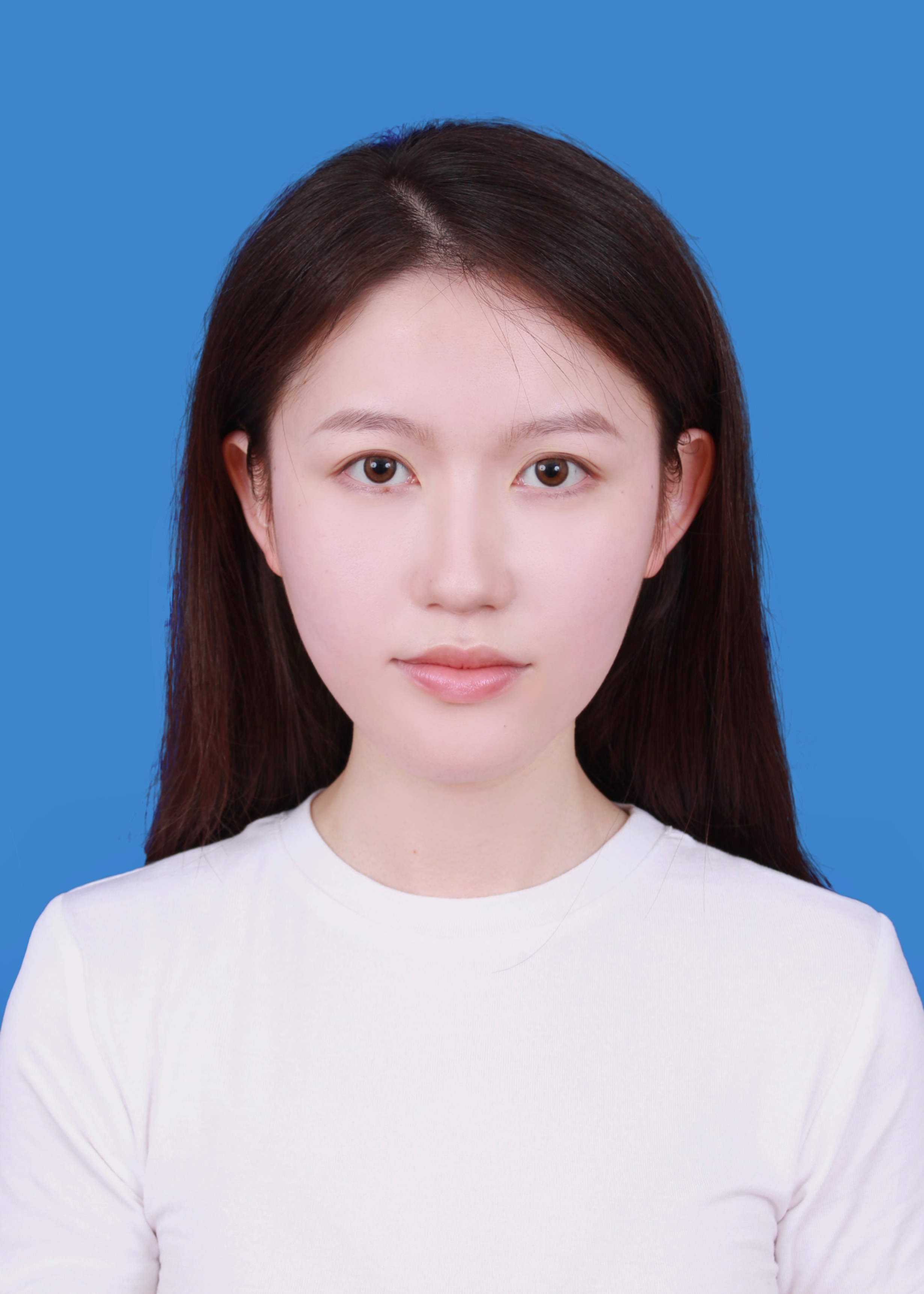}}]{Yunshuang Nie} received the B.E. degree in Sun Yat-sen University, Shenzhen, China, in 2023. She is currently working toward the M.E. in the school of intelligent systems engineering of Sun Yat-sen University. Her current research interests include vision-and-language understanding and embodied AI.
\end{IEEEbiography}

\begin{IEEEbiography}[{\includegraphics[width=1in,height=1.25in,clip,keepaspectratio]{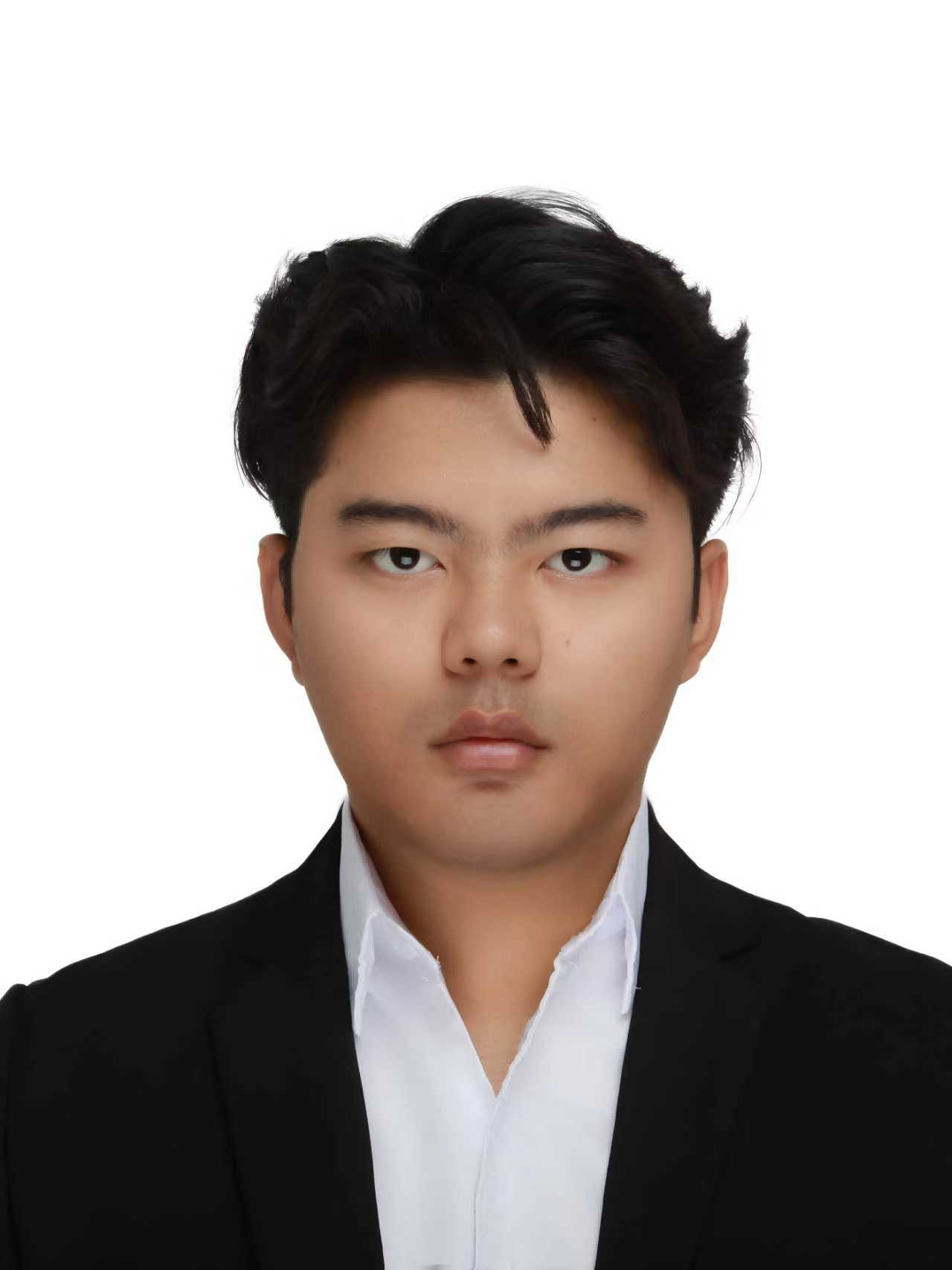}}]{Khun Loun Zai} received the B.E. degree in Sun Yat-sen University, Shenzhen, China, in 2025. He is an M.E. candidate in Computer Science at Peking University. His research focuses on vision-and-language understanding and embodied AI.
\end{IEEEbiography}

\begin{IEEEbiography}[{\includegraphics[width=1in,height=1.25in,clip,keepaspectratio]{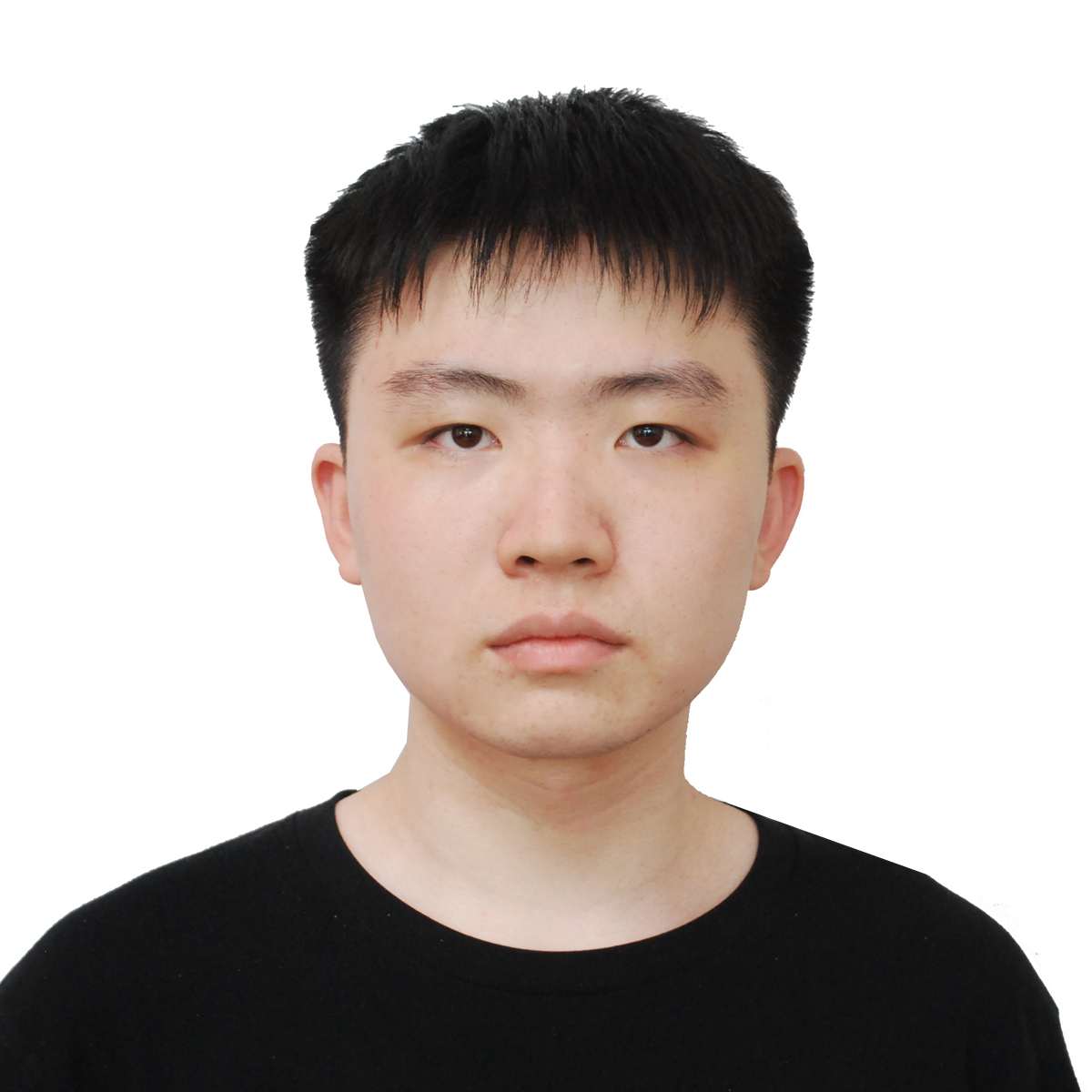}}]{Ziming Wei} received the B.E. degree in intelligence science and technology from Sun Yat-sen University in 2024. He is currently pursuing the M.S. degree with the school of intelligent systems engineering of Sun Yat-sen University. His current research interests include multi-modal understanding, learning and data generation, embodied AI and spatial intelligence.
\end{IEEEbiography}

\begin{IEEEbiography}[{\includegraphics[width=1in,height=1.25in,clip,keepaspectratio]{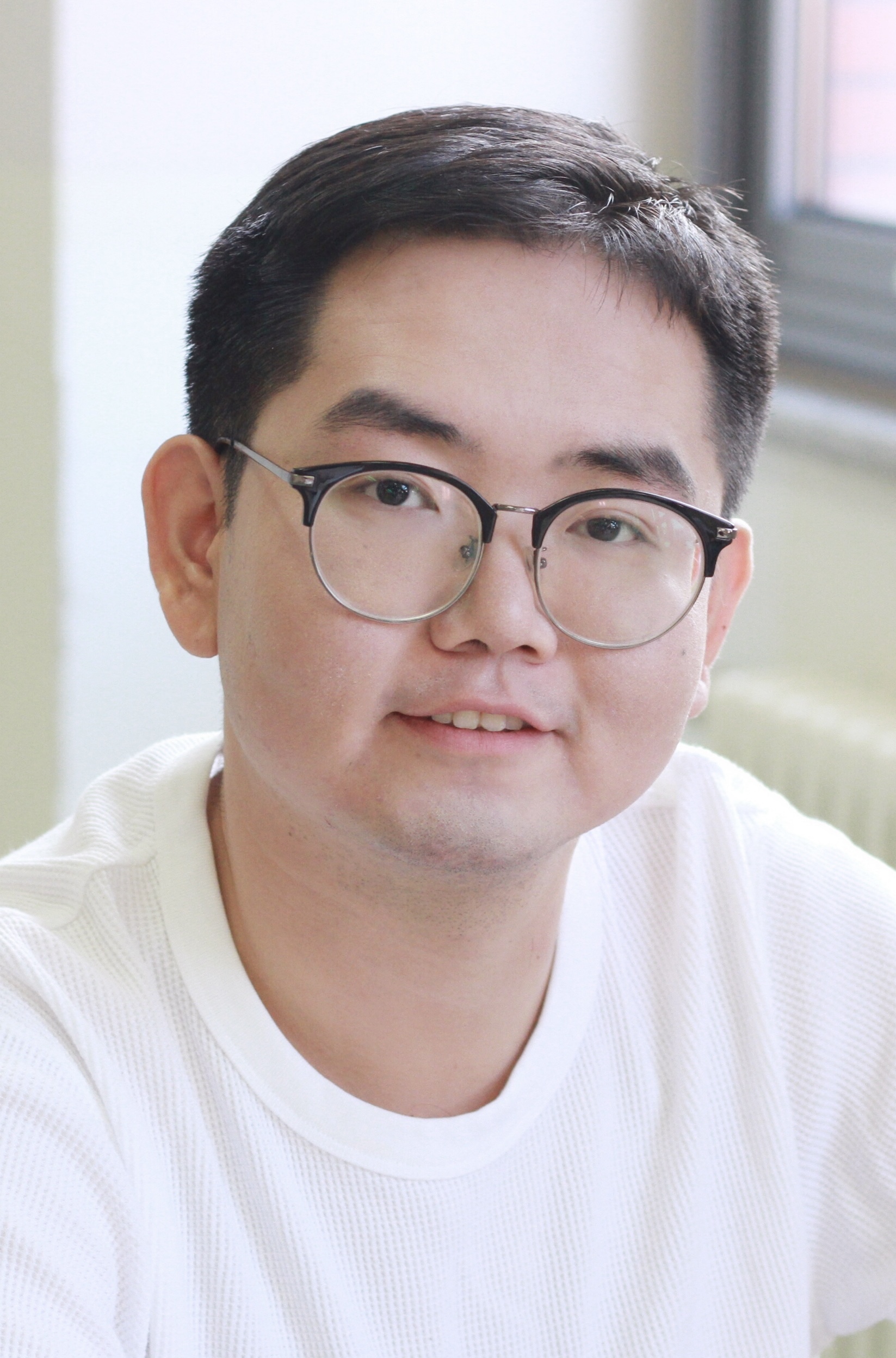}}]{Mingfei Han} is currently a postdoctoral associate at Mohamed Bin Zayed University of Artificial Intelligence. He obtained his Ph.D. degree from University of Technology Sydney. He received the B.Eng. degree from Nankai University and the M.Eng. degree from University of Chinese Academy of Sciences. His research interests lie in computer vision and machine learning, with a particular emphasis on large vision-language models, video object perception and their applications in robotics.
\end{IEEEbiography}

\begin{IEEEbiography}[{\includegraphics[width=1in,height=1.25in,clip,keepaspectratio]{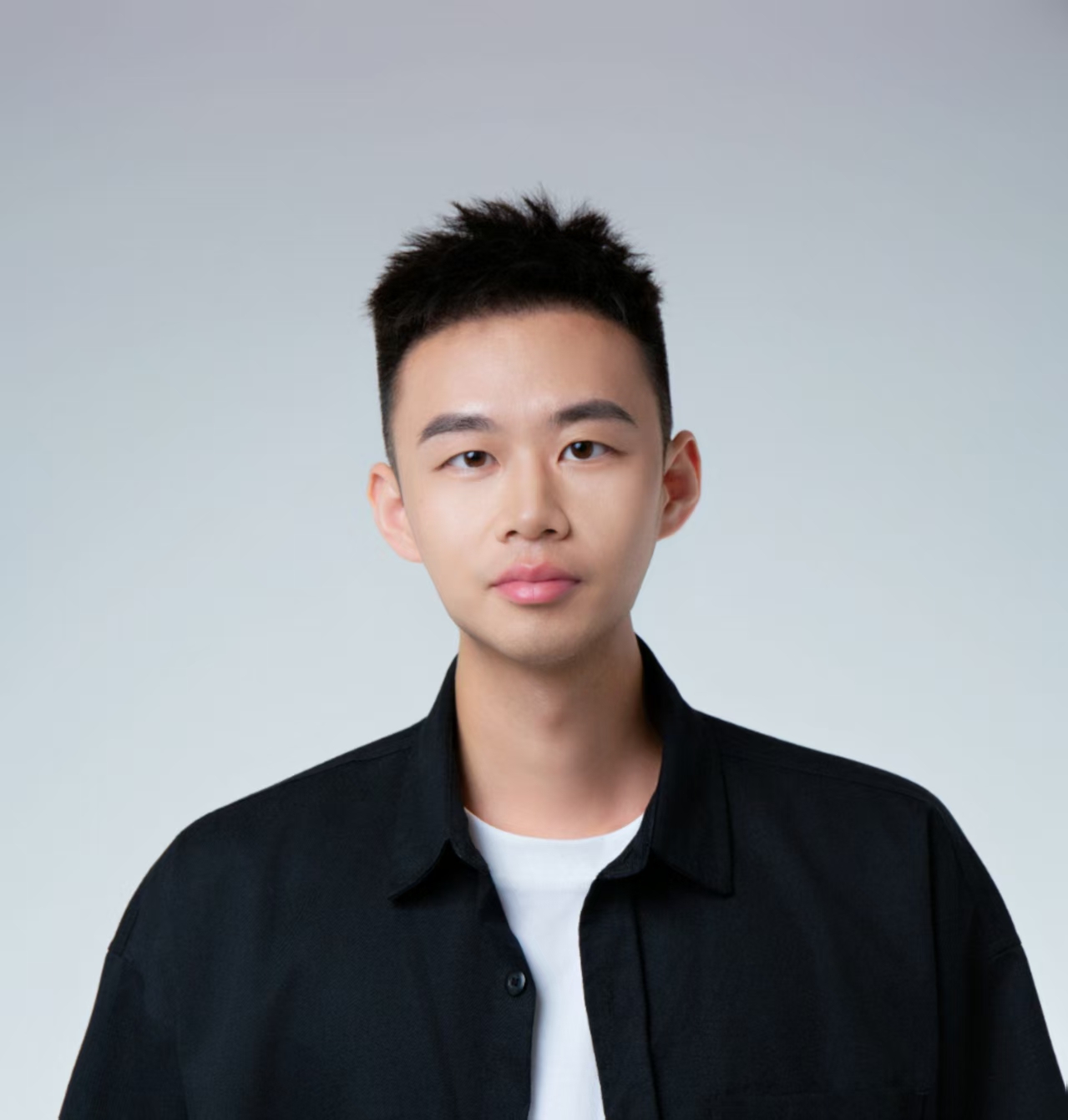}}]{Rongtao Xu} is currently a Postdoctoral Researcher at Mohamed bin Zayed University of Artificial Intelligence (MBZUAI), advised by Prof. Xiaodan Liang. He received the B.S. degree in information and computing science from Huazhong University of Science and Technology, China, in July 2019. From September 2019 to 2024, he is a Ph.D student majoring in the State Key Laboratory of Multimodal Artificial Intelligence Systems, Institute of Automation, Chinese Academy of Sciences and School of Artificial Intelligence, University of Chinese Academy of Sciences. His research interests include embodied AI, multimodal learning and robotic vision.
\end{IEEEbiography}

\begin{IEEEbiography}
[{\includegraphics[width=1in,height=1.25in, clip,keepaspectratio]{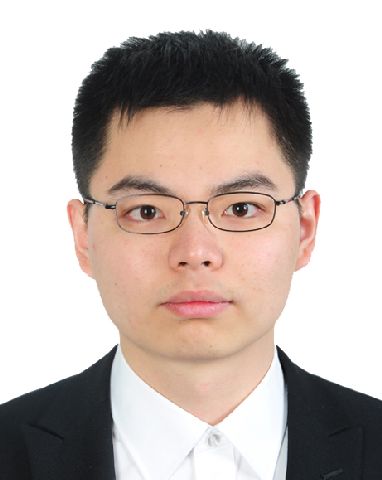}}]{Minzhe Niu} is currently a researcher with Yinwang Intelligent Technology Co., Ltd. He received his B.E. and M.E. degrees in Shanghai Jiao Tong University. His research interest includes multi-modality learning, autonomous driving and machine learning.
\end{IEEEbiography}

\begin{IEEEbiography}
[{\includegraphics[width=1in,height=1.25in, clip,keepaspectratio]{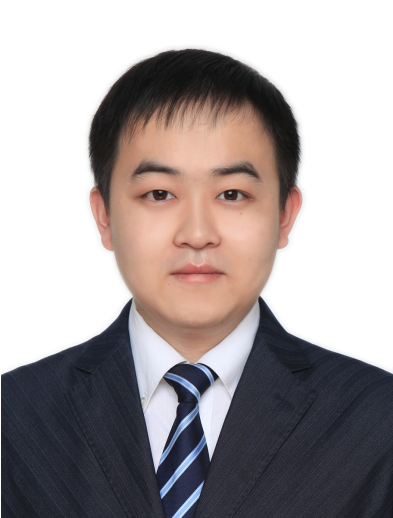}}]{Jianhua Han} received the Bachelor Degree in 2016 and Master Degree in 2019 from Shanghai Jiao Tong University, China. He is currently a researcher with Yinwang Intelligent Technology Co., Ltd. His research interests lie primarily in deep learning and computer vision.
\end{IEEEbiography}

\begin{IEEEbiography}[{\includegraphics[width=1in,height=1.25in,clip,keepaspectratio]{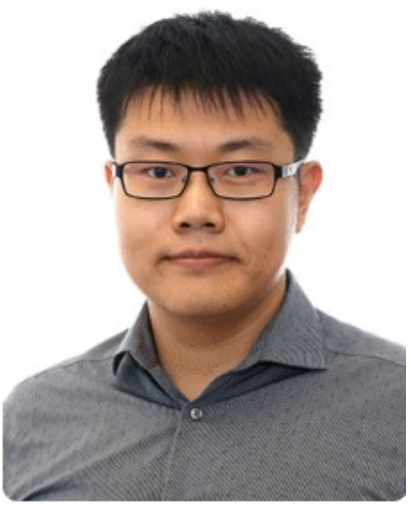}}]{Hanwang Zhang} received the BEng.(Hons.) degree in computer science from Zhejiang University,
Hangzhou, China, in 2009, and a Ph.D. degree in
computer science from the National University of
Singapore (NUS), Singapore, in 2014. He is currently an associate professor with Nanyang Technological University, Singapore. His research interests
include developing multi-media and computer vision
techniques for efficient search and recognition of
visual content. He received the Best Demo RunnerUp Award in ACM MM 2012 and the Best Student
Paper Award in ACM MM 2013. He was the recipient of the Best Ph.D.
Thesis Award of the School of Computing, NUS, 2014.
\end{IEEEbiography}

\begin{IEEEbiography}[{\includegraphics[width=1in,height=1.25in,clip,keepaspectratio]{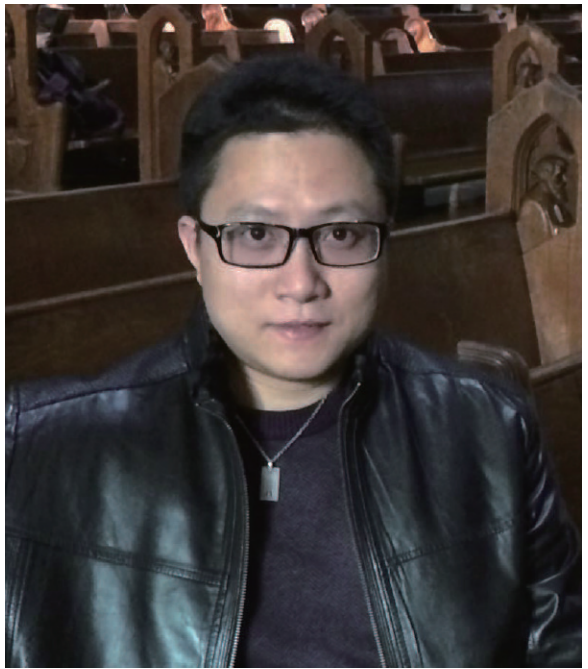}}]{Liang Lin} (Fellow, IEEE) is a Full Professor of
computer science at Sun Yat-sen University. He
served as the Executive Director and Distinguished
Scientist of SenseTime Group from 2016 to 2018,
leading the R\&D teams for cutting-edge technology
transferring. He has authored or co-authored more
than 200 papers in leading academic journals and
conferences, and his papers have been cited by more
than 30,000 times. He is an associate editor of IEEE
Trans.Neural Networks and Learning Systems and
IEEE Trans. Multimedia, and served as Area Chairs
for numerous conferences such as CVPR, ICCV, SIGKDD and AAAI. He is
the recipient of numerous awards and honors including Wu Wen-Jun Artificial
Intelligence Award, the First Prize of China Society of Image and Graphics,
ICCV Best Paper Nomination in 2019, Annual Best Paper Award by Pattern
Recognition (Elsevier) in 2018, Best Paper Dimond Award in IEEE ICME
2017, Google Faculty Award in 2012. His supervised PhD students received
ACM China Doctoral Dissertation Award, CCF Best Doctoral Dissertation
and CAAI Best Doctoral Dissertation. He is a Fellow of IEEE/IAPR/IET.
\end{IEEEbiography}

\begin{IEEEbiography}[{\includegraphics[width=1in,height=1.25in,clip,keepaspectratio]{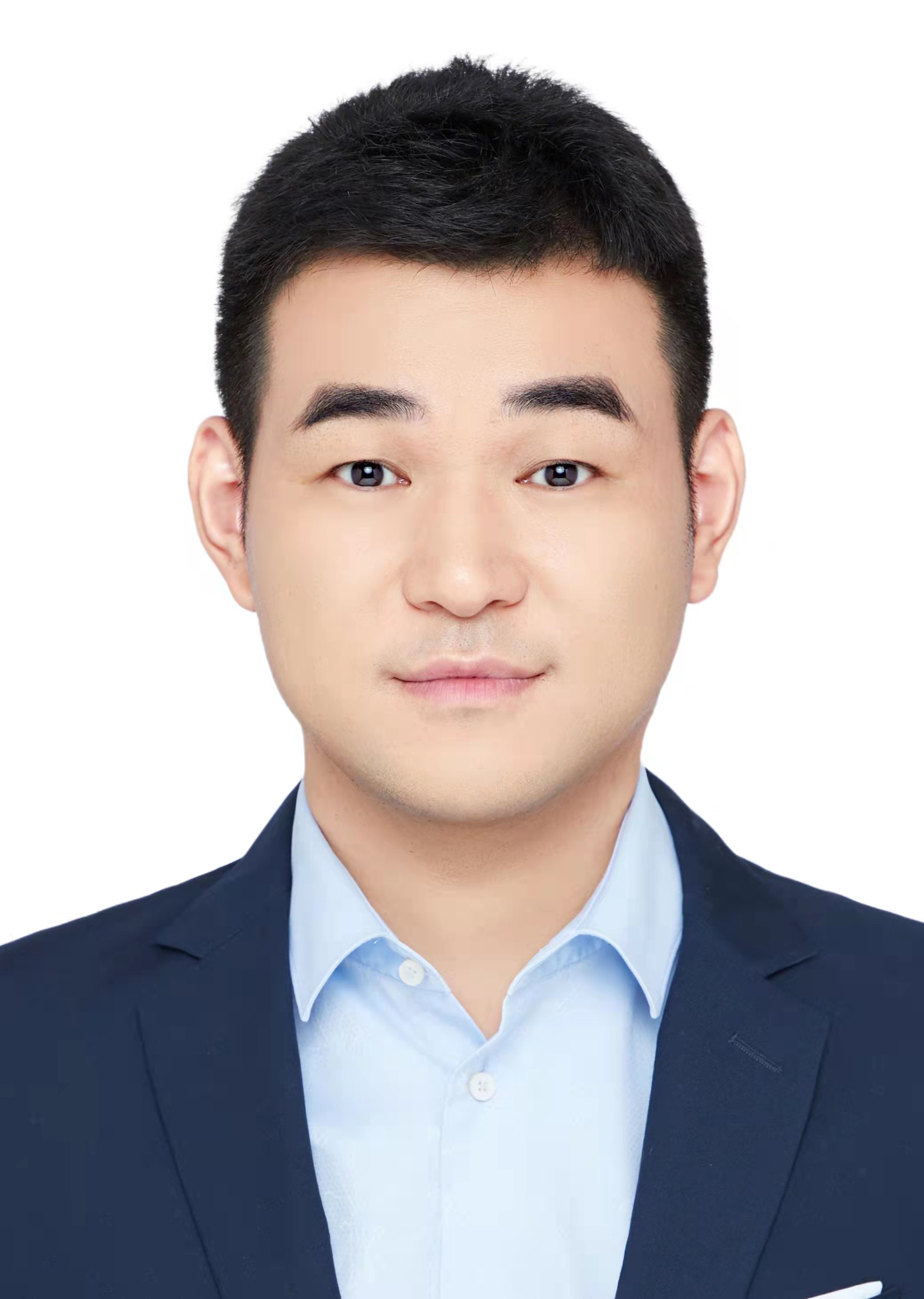}}]{Bokui Chen} received the Ph.D. degree from University of Science and Technology of China in 2013. He is currently an Assistant Professor at Tsinghua Shenzhen International Graduate School, Tsinghua University, China. His research interests include intelligent transportation systems and artificial intelligence.
\end{IEEEbiography}

\begin{IEEEbiography}[{\includegraphics[width=1in,height=1.25in,clip,keepaspectratio]{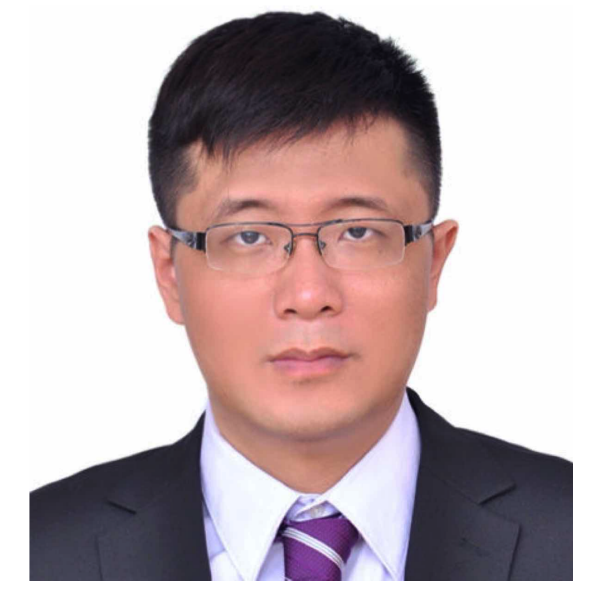}}]{Cewu Lu} is a Professor at Shanghai Jiao Tong
University (SJTU). Before he joined SJTU, he was a research fellow at Stanford University,
working under Prof. Fei-Fei Li and Prof. Leonidas
J. Guibas. He was a Research Assistant Professor at Hong Kong University of Science and
Technology with Prof. Chi Keung Tang. He got
his Ph.D. degree from the Chinese University of
Hong Kong, supervised by Prof. Jiaya Jia. He is one of the core technique members in Stanford Toyota autonomous car project. He serves as
an associate editor for journal CVPR and reviewer for journal TPAMI
and IJCV. His research interests fall mainly in computer vision, deep
learning, deep reinforcement learning and robotics vision.
\end{IEEEbiography}

\begin{IEEEbiography}[{\includegraphics[width=1in,height=1.25in,clip,keepaspectratio]{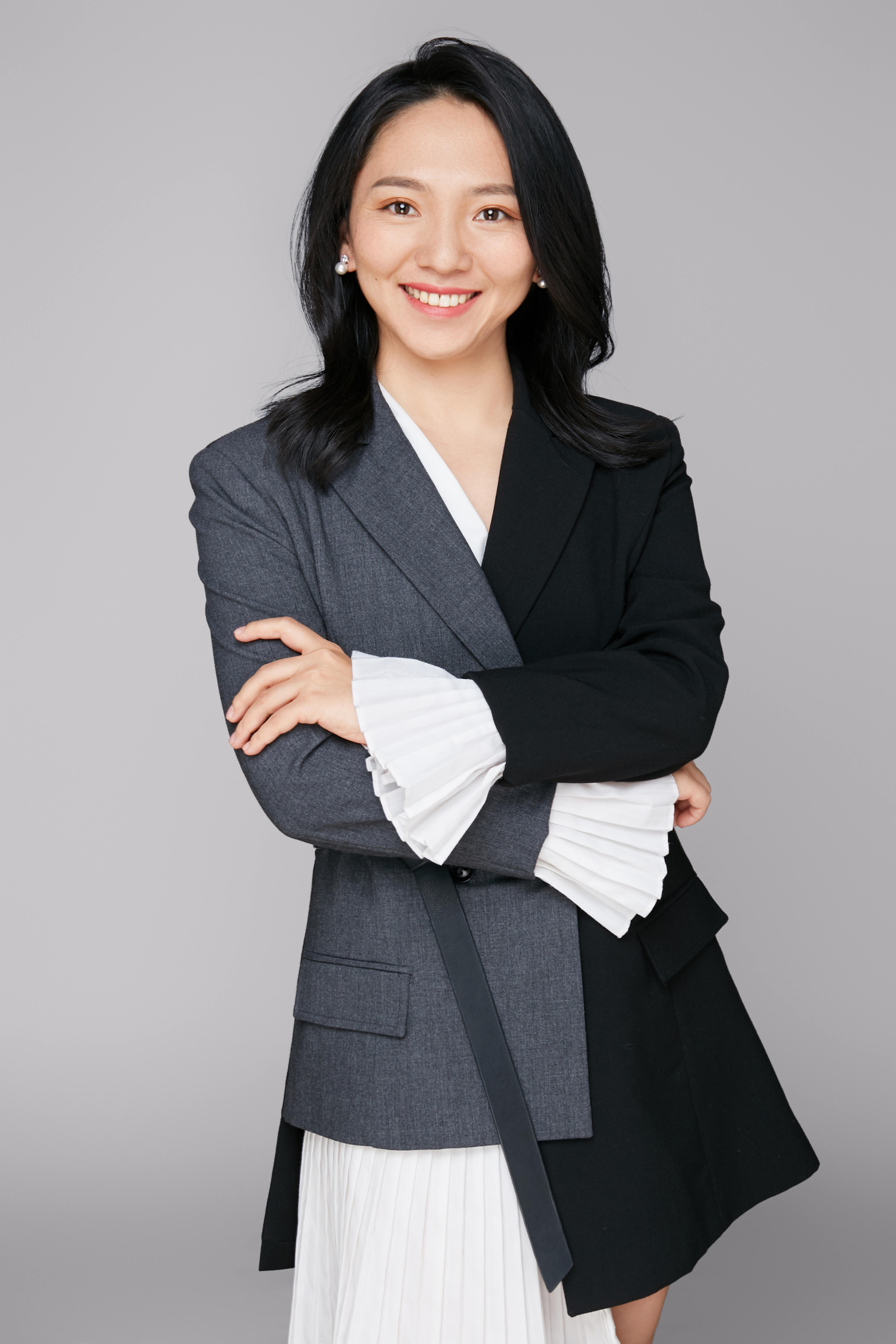}}]{Xiaodan Liang} 
received the Ph.D. degree from
Sun Yat-sen University, Shenzhen, China, in 2016,
advised by Liang Lin.
She was a Post-Doctoral Researcher with the
Machine Learning Department, Carnegie Mellon
University, Pittsburgh, PA, USA, from 2016 to 2018,
working with Prof. Eric Xing. She is currently
an Associate Professor with Sun Yat-sen University. She has published several cutting-edge projects
on human-related analysis, including human parsing, pedestrian detection and instance segmentation,
2D/3D human pose estimation, and activity recognition.
\end{IEEEbiography}

\vfill

\end{document}